\documentclass[fleqn,10pt]{wlscirep}
\usepackage[utf8]{inputenc}
\usepackage[T1]{fontenc}
\usepackage{algorithm}
\usepackage{algorithmic}
\usepackage{caption}
\usepackage{subcaption}
\usepackage{array,booktabs,arydshln,xcolor}
\usepackage{adjustbox}
\usepackage{tikz}
\usetikzlibrary{shapes.geometric,  automata, positioning, arrows.meta, matrix}
\tikzstyle{startstop} = [rectangle, rounded corners, minimum width=3cm, minimum height=1cm,text centered, draw=black, fill=red!30]
\tikzstyle{io} = [trapezium, trapezium left angle=70, trapezium right angle=110, minimum width=3cm, minimum height=1cm, text centered, draw=black, fill=cyan!30]
\tikzstyle{process} = [rectangle, minimum width=3cm, minimum height=1cm, text centered, draw=black, fill=orange!30]
\tikzset{decision/.style={diamond, minimum width=3cm, minimum height=1cm, text centered, draw=black, fill=green!30}}
\tikzset{arrow/.style={thick,->,>=stealth}}
\title{Learning to Play Chess from Textbooks (LEAP):\\ a Corpus for Evaluating Chess Moves based on Sentiment Analysis}

\author[1,$\dag$]{Haifa Alrdahi}
\author[1,$\dag$]{Riza Batista-Navarro}

\affil[1]{Department of Computer Science, University of Manchester, Manchester, M13 9PL, United Kingdom}

\affil[*]{corresponding author(s): Haifa Alrdahi (haifa.alrdahi@manchester.ac.uk, haifa.alrdahi@gmail.com)}

\affil[$\dag$]{these authors contributed equally to this work}

\begin{abstract}

Learning chess strategies has been investigated widely, with most studies focussing on learning from previous games using search algorithms. Chess textbooks encapsulate grandmaster knowledge, explain playing strategies and require a smaller search space compared to traditional chess agents. This paper examines chess textbooks as a new knowledge source for enabling machines to learn how to play chess---a resource that has not been explored previously. We developed the LEAP corpus, a first and new heterogeneous dataset with structured (chess move notations and board states) and unstructured data (textual descriptions) collected from a chess textbook containing 1164 sentences discussing strategic moves from 91 games. We firstly labelled the sentences based on their relevance, i.e., whether they are discussing a move. Each relevant sentence was then labelled according to its sentiment towards the described move. We performed empirical experiments that assess the performance of various transformer-based baseline models for sentiment analysis. Our results demonstrate the feasibility of employing transformer-based sentiment analysis models for evaluating chess moves, with the best performing model obtaining a weighted micro ${F}_{1}$ score of 68\%. Finally, we synthesised the LEAP corpus to create a larger dataset, which can be used as a solution to the limited textual resource in the chess domain.

\end{abstract}
\begin{document}

\flushbottom
\maketitle

\thispagestyle{empty}


\section*{Background \& Summary}

Chess is a stochastic environment controlled by the rules of playing, which are simple to comprehend, yet it is challenging to make accurate decisions in the game. Hence, chess lends itself well to the development of an artificial intelligence (AI) system that simulates real-life problems, such as decision-making \cite{yannakakis2018artificial}. Meanwhile, chess grandmasters had produced, and continue to produce chess-teaching textbooks, in unstructured data format, for sharing their practical knowledge on strategies. Chess-teaching textbooks comprise a substantial knowledge source, out of many others (e.g., game commentaries), that chess players continuously use to grasp knowledge of strategies and tactics and improve their skills \cite{ross2006expert}. Over the years, little effort has been put to explore chess knowledge from unstructured data sources. Many machine learning algorithms for playing chess have thus far overlooked the potential of obtaining knowledge from chess-teaching textbooks. Instead, knowledge is typically obtained from databases of chess moves, such as \textit{DeepChess} \cite{david2016deepchess}. Such an approach is reliant on large curated structured datasets of chess moves capturing strategies of expert players, such as \textit{Chess Database} \cite{Schaigorodsky2016}. The production of such datasets is often laborious and time-consuming \cite{nechepurenko2019comparing} and requires intensive work for the interpretability and explainability in the decision-making process of the move \cite{andrulis2020domain, anjomshoae2019explainable}. Additionally, using brute-force approaches, such as AlphaZero \cite{silver2017mastering}, to obtain high performance requires expensive computational resources including advanced hardware requirements that might not be accessible or obtainable for research \cite{nechepurenko2019comparing,kamlish2019sentimate}.

Processing knowledge from unstructured data is intricate due to knowledge accessibility, the language's complex nature \cite{sanchez2021unleashing,feng2018extracting}, and understanding of the domain environment expressed in natural language. However, various studies in different domains have shown that using unstructured data for an alternative approach to overcome some limitations of using brute-force approaches only. Previous work focused mainly on extracting actions from short direct instruction sentences. This alternative approach has led to the improvement of the AI systems performance. Information extraction (IE) methods for identifying entities and relations from the \textit{Civilization II} instructional manual were integrated into a Monte Carlo Tree Search (MCTS) model to learn how to play the game \cite{branavan2012learning}. The natural language-based model demonstrated improved performance by 33.9\% of the time compared to an AI agent without knowledge of the instructional manuals. In the video-games domain, a work to analyse Steam platform reviews was performed for identifying game features, sentiment towards players and identifying spam reviews \cite{jeffrey2020wisdom}. Similarly, Events Extraction (EE) from unstructured data was used for continuously updating a stochastic model that performs operational processes in uncertain and changing environments, such as evacuation routes recommended by humans using the Twitter platform \cite{paterson2019using}. An LSTM-RNN model was developed to translate instruction sentences of the current state into action sequences for autonomous agents \cite{mei2016listen}.
The recent advances in context representation and language analysis led to the development of state-of-the-art pre-trained language models. Such models were pre-trained on a large corpus and then fine-tuned for downstream tasks in different domains. For example, sentiment analysis of chess commentaries representation using an LSTM model with BERT embeddings \cite{kamlish2019sentimate} improved the alpha-beta chess evaluation function, which won 81\% of 100 games against random and DeepChess \cite{david2016deepchess} systems. Recently, GPT-2 model was trained on 2.8 million chess games in Portable Game Notation (PGN) to predict the next move \cite{noever2020chess}. A different approach of analysing how GPT-2 learns chess playing rules, store chess knowledge and generate a move was studied in \cite{stockl2021watching}, which took into consideration different model sizes, the size of the training and the number of correctly generated moves. An LSTM model was used to generate chess moves commentaries benchmark that are comparable to human commentaries in terms of grammar and language 
\cite{jhamtani2018learning}.
Furthermore, DistilBERT \cite{sanh2019distilbert} was used to encode state and action representations of the text game, then fed them into Reinforcement Learning (RL) agent, which achieved 44.7 a new state-of-the-art result in the interactive textual game \textit{Zork1}. 

Nonetheless, the usability and the benefit of free unstructured data approach, such as textbooks, has not been explored before and there is almost no previous research in the literature to explore this approach in board games domain, specifically in chess domain. In this paper, we are introducing a new task in the game playing domain that is underpinned by Natural Language Processing (NLP): sentiment analysis for unlocking the otherwise hidden knowledge of chess master players from unstructured data.
The contributions of this paper is four folds; firstly, we introduce LEAP, LEArning to Play chess, a new heterogeneous corpus collected from Chess-teaching Textbooks in believe that it can be used to aid natural language-based chess agents to evaluate moves from heterogeneous chess knowledge sources. The corpus contains two data types; labelled sentences that discuss the game's moves, "unstructured data", and the full game's move with their board states "structured data". We believe that the latter data format is necessary to link the chess context, expressed in sentences, to its equivalent environment represented in board states. 
Secondly, We introduce two types of data annotations (in the above-mentioned corpus): (1) \emph{relevancy} labels indicating whether a sentence is discussing a move, and (2) \emph{sentiment} labels, i.e., whether a move is considered as good or bad, where labels are cast as an evaluation task of the move. 
Thirdly, we demonstrate by empirical evaluations the usability and characteristics of the corpus using state-of-the-art transformer-based models for the two classification tasks. We report the performance of various transformer models as baselines, discuss further improvements and propose approaches for move evaluation. Finally, we contribute to the limited unstructured resources problem in the chess domain by synthesising the LEAP corpus to create a larger dataset, and show by empirical experiments the quality and usability of the dataset for training language models.

\section*{Methods}
\subsection*{Challenges}
\begin{figure*}[h]
\centering
   \centering\includegraphics[width=17cm, height=11cm]{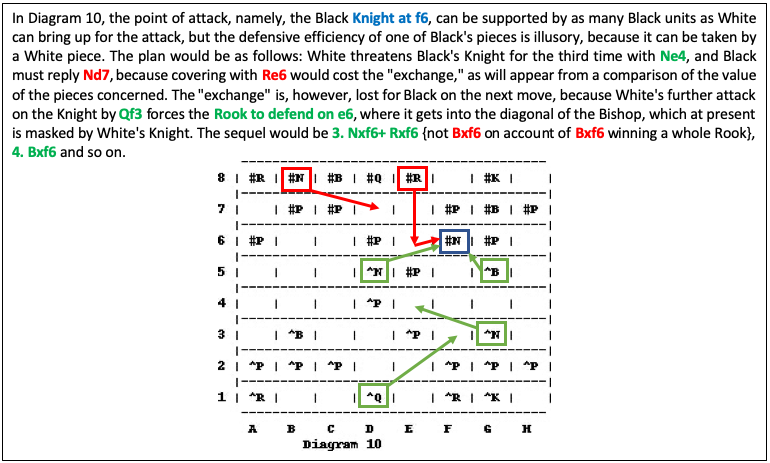}

   \caption{The example is taken from a chess-teaching textbook; Chess Strategy by Edward Lasker (1918)\url{https://www.gutenberg.org/ebooks/5614}, an chess player on an estimated international master level with an estimated ELO rating of 2489. The moves written style in the textbook have been edited here from Descriptive notation to Algebraic notation for readability purposes. Green colour refers to White moves, Red colour refers to Black moves, Blue colour is the piece under attack, the colours were added for illustration purpose.}
   \label{figure:1}
\end{figure*}

A textbook is a knowledge acquisition and learning source, but there are several challenges that come with mining chess-teaching textbooks. Firstly, chess playing \textit{\textbf{Illustration}} is not limited to text, but to board state as well, an example of textbook's paragraph is shown in Figure \ref{figure:1}. It is difficult to deduct why a move described in text is necessary for a player without visualisation of the board state. Also, showing a board state diagram and a list of moves only is not useful without explaining why these moves are chosen and have been played in this order. Thus, the board state diagrams help the reader visualising the environment and understand the connection with the described moves in the text. Chess-teaching textbooks provide access to both data formats; the textual description of the states and the legal move(s) to be considered in these states, and also refers to diagrams of the board state (environment).

The second challenge in mining such a knowledge source is \textit{\textbf{Dependency}}. Chess game consists of a sequence of moves, where each played move creates a new board state, changes the set of plausible, legal moves, and affects board evaluation for the players. An example sentence is in Figure \ref{figure:1}; "\textit{White's further attack on the Knight by Qf3 forces the Rook to defend on K3}". This means that Black's move "\textit{Rook to defend on K3}" is favoured only if White played "\textit{Qf3}". Hence, the decision-making process with respect to moves, requires recognising the move's effect on the board state after the move has been made, which eventually creates a dependency between the current board state and the moves being considered. 

\textit{\textbf{Incompleteness or Implicitness}} of information required for decision-making is a challenge presented in textbooks. For example, the move to be played in the following sentence, taken from Chess Strategy by Edward Lasker (1918)\footnote{\url{https://www.gutenberg.org/ebooks/5614}}, is missing and not described explicitly in text: \textit{``Black would appear to have sufficient protection available, with his Knight and Bishop.''}. The move with the Knight or the Bishop is implicitly favoured. However, it is not straightforward for a chess agent to identify this move, typically requiring the use of search algorithms.

The fourth challenge is related to \textit{\textbf{Recommendation of Conflicting Moves}}: Some sentences describe a move, or a sequence of moves, using counterfactual statements which could confuse a chess agent, leading to the extraction of inaccurate moves and hence incorrect actions. An example of counterfactual statement can be seen in the following sentences: \textit{``If White had only a Bishop or a Knight Additionally to the King he could never mate Black, for neither Bishop nor Knight can attack the King and at the same time control a square adjacent to the King. This, however, is at least necessary to force the mate, even in the most unfavourable position of the King, that is, in the corner.''}. At first, the author explains that the move "\textit{mate Black King}" by the White Bishop or a Knight is discouraged in this game state, but also suggests in the second sentence that this move is necessary to be played.

The last challenge we observed is \textit{\textbf{Poor Formatting}}: Information is not always organised or presented in a consistent manner. Textbooks can describe special moves in natural language, such as ``\textit{pawn is promoted to a queen}'', or in chess notation such as Standard Algebraic Notation (SAN) ``\textit{Pe8=Q}'', or in both ``\textit{Qc8, white was just promoted to a queen, giving mate}''.

Some of the challenges involved in mining chess-teaching textbooks are also pertinent in other problems that require decision-making, such as in computer vision \cite{schwartz2019natural}, science literature understanding \cite{mysore2019materials}, medical domains \cite{hogenboom2016survey} and interacting with robotics and human-machine interaction using natural language \cite{alomari2017natural}.  

\subsection*{Case study context}
The goal of evaluating chess moves is to find an optimal move $Move_{o}$ for a state $State_{s}$. Since 1949, Claude Shannon designed the evaluation formula in a heuristic structure to determine the relative value of a move $Move_{m}$ by measuring the score of a board state $State_{s}$ after playing $Move_{m}$. There are different features to consider while evaluating a move $Move_{m}$ heuristically, such as the game status e.g. middle-game or end-game, the pieces-positions values and the king safety based on its position. Currently, these heuristics are embedded in the chess engines with search algorithms to evaluate chess moves. In general, the search algorithms are based on tree structure, where the nodes of the tree represent all possible future states of each legal move for the player, the depth level represents the result state of playing a legal move, e.g. $Move_{m}$, and the edges represents the transition from a state $State_{s}$ to state $State_{s+1}$ by playing $Move_{m}$. Alpha-beta pruning search algorithm has been widely used to evaluate chess moves, which is a directed graph based on Min-Max algorithm. To determine which move should be played at $State_{s}$, the Alpha-beta algorithm first removes (prunes) the unused tree branch, then visits all the remaining nodes (moves) and evaluates the board state score after playing each move. Finally, Alpha-beta backtracks and selects the optimal move $Move_{o}$ for the $State_{s}$. The second search algorithm is Monte Carlo Tree search algorithm which is based on simulations games, where each simulation game is the result of a randomly selected moves. Finally the optimal move $Move_{o}$ is selected based on the game that achieved highest result.

\begin{figure*}[h]
   \centering
\resizebox{\textwidth}{!}{
\begin{tikzpicture}[node distance=2.5cm, auto]

\node (in1) [io, text width=4.5cm, align=center] {"After the first few moves, we arrive at the following position, which may be reached thus:\\
3. Bb5 d6        4. d4 Bd7\\
5. Nc3 Nc6      6. O-O Be7\\
7. Re1 \textcolor{red}{exd4}     8. Nxd4 O-O.\\ 
The \textcolor{red}{exchange} on the seventh move is compulsory, because the loss of a pawn after Bxc6 is in effect threatened, now that the White e Pawn is supported by the Rook"
};
\node (pro1) [process, right of=in1, xshift=5cm, text width=1.7cm, align=center] {Move Extraction}; 
\node[inner sep=0pt] (fig1) [right of=pro1, xshift=3cm] {\includegraphics[width=4cm]{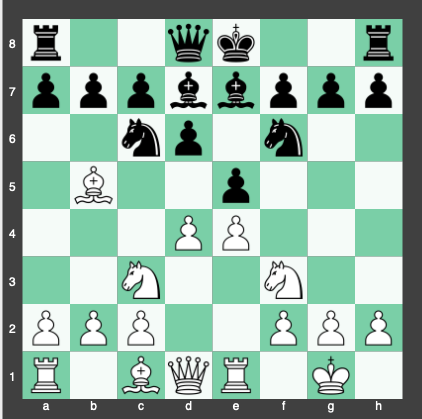}};
\node (in2) [io, below of=pro1, text width=3cm] {“The exchange on the seventh move” (exd4)};
\node (pro2) [process,  below of=in2,  text width=1.7cm, align=center] {Move Validation};
\node (pro3) [process, xshift=3cm, right of=pro2,  text width=1.1cm, align=center] {Chess Engine};
\node (pro4) [process,  below of=pro2, yshift=-0.5cm, text width=1.7cm, align=center] {Move Evaluation};
\node (pro5) [process,  below of=in1, yshift=-2.5cm, text width=1.7cm, align=center] {Sentiment Analysis};
\node (in3) [io, left of=pro4, xshift=-5cm, yshift=0cm, text width=3cm] {Sentiment: Positive, “is compulsory”};
\node[inner sep=0pt] (fig2) [below of=pro3, yshift=-3cm, xshift=0cm] {\includegraphics[width=4cm]{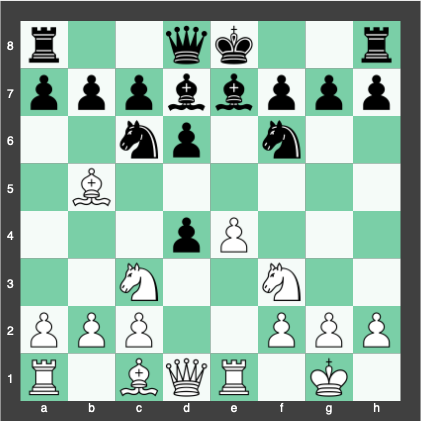}};
\node (pro6) [process,  below of=pro4, yshift=-0.5cm, text width=1.7cm, align=center] {Play exd4};

\draw [arrow] (in1) -- (pro1);
\draw[thick] (15,0) -- (16,0);
\draw[thick] (16,0) -- (16,-8);
\draw[thick] (16,-8) -- (13,-8);
\draw [arrow] (fig1) -- (pro3);
\draw [arrow] (fig1) -- (pro1);
\draw [arrow] (pro1) -- (in2);
\draw [arrow] (in2) -- (pro2);
\draw [thick,latex-latex] (pro2) -- node[anchor=south] {Valid Move} (pro3);
\draw [arrow] (pro2) -- (pro4);
\draw [arrow] (in1) -- (pro5);
\draw [latex-latex] (pro3) |- (pro4);
\draw [arrow] (pro5) -- (in3);
\draw [arrow] (in3) -- (pro4);
\draw [arrow] (fig2) |- (pro4);
\draw [arrow] (pro4) -- (pro6);


\end{tikzpicture}}
   \caption{Example sentence from a chess-teaching textbook shows an analysis skeleton for a natural language-based chess agent to process the text, identify the discussed moves and evaluate them accordingly.}
   \label{figure:2}
\end{figure*}
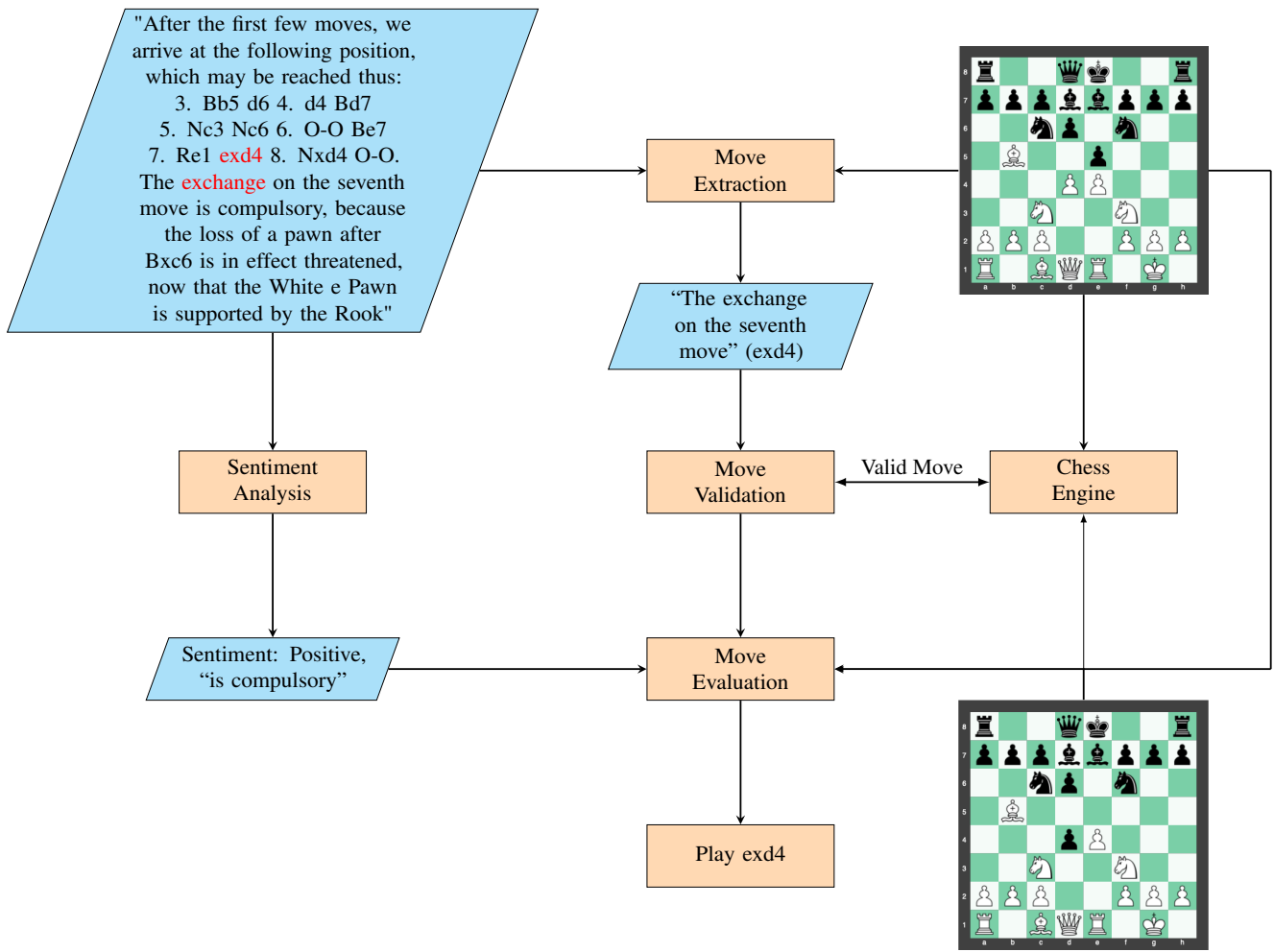

The scope of this work is to extract the evaluation function of the moves from the unstructured data "natural language" description in chess-teaching textbooks that evaluates a move. Our scope is different from previous work in the literature by mining this rich content instead of itemised instructions, such as in \textit{Civilization II} instructional manual \cite{branavan2012learning}. In this on-going work, our first step is aiming to bridge the gap for a chess agent understanding of this description during decision-making processes. Figure \ref{figure:2} shows an example sentence and how it can be analysed and understood by a chess agent. A textbook sentence usually is a descriptive rather than a direct instruction of an action or move. The example sentence explains \emph{why} Black needs to play the move "\textit{the exchange on the seventh move}", rather than directly instructing the chess agent to play the move. To train a natural language-based chess agent understanding such an action, we need information extraction methods, possibly Named-Entity Recognition (NER), to identify the player and possible moves from the discussed moves $Moves=(Move_{1}, ... , Move_{n})$, where ${n}$ is the number of moves discussed in a sentence. In addition, it is necessary to integrate the current board state $State_{s}$ with moves extracted from the sentences to identify which of the discussed moves is a possible one. Hence, we formulate the information as a tuple of each extracted move: \[ Board\;Evaluation (Player, Move_{i}, State_{s}) \] 
the tuple is sent to a chess engine to validate the move legally at this specific board state $State_{s}$, then evaluate it using search algorithms (e.g., alpha-beta pruning).
However, following the related work in analysing chess commentaries, we hypothesis that it is possible to infer the move evaluation by analysing the text description of the move's effect using sentiment analysis. The outcome of a move is usually described in textbook to either have a negative, positive or neutral effect on the player or sometimes to the opponent. The example sentence in Figure \ref{figure:2} highlights that the outcome of playing the move "\textit{the exchange, exd4}" is positive for Black in this turn "\textit{is compulsory}" by explaining that a Black Pawn will be lost later. In other words, if the move "\textit{the exchange, exd4}" is played it will increase the black score and reduce the white score by losing his pawn. Then it goes on to explain the necessity of the move "\textit{is compulsory}", because the black pawn will be lost in the next move by Nxd4. This will lead to lower back the Black player score and increase the White player score. Hence, the move "\textit{the exchange, exd4}" helps Black player to maintain his score for another turn, and without the move his score would decrease in the second turn. A human player can easily interpret the discussed moves and their effects, filter and choose the move with a positive impact while considering the current board at the same time. For a natural language-based chess agent, we can cast the move evaluation process as the analysis of a sentiment towards the player. A positive sentiment indicates that playing the move would likely have a positive effect on the player. This can be explicitly stated, such as "it is best to play move X", or implicitly derived as in the example in Figure \ref{figure:2}. A negative sentiment indicates a negative effect on the player: either explicitly, such as "it is best to avoid playing move X", or implicitly as in "playing move X will help opponent player to progress".

\subsection*{Dataset sources}
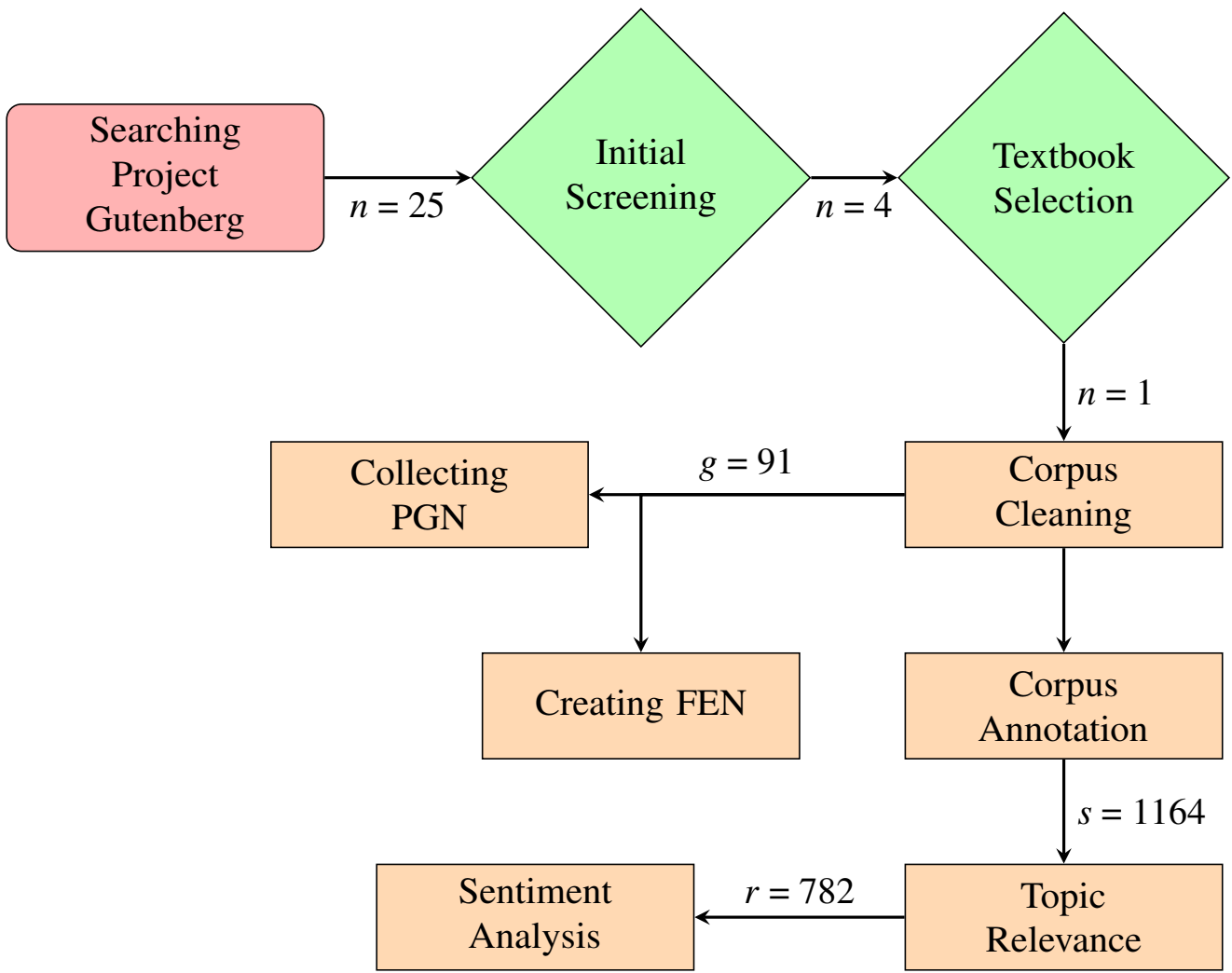
\begin{figure*}[h]
   \centering
\resizebox{\textwidth}{!}{
\begin{tikzpicture}[>=stealth, node distance=2cm, auto]
\node (start) [startstop, text width=1.7cm, align=center] {Searching Project Gutenberg};
\node (dec1) [decision, right of=start, xshift=2.5cm, text width=2cm, align=center] {Initial Screening};
\draw [arrow] (start) -- node[anchor=north] {$n$ = 25} (dec1);
\node (dec2) [decision, right of=dec1, xshift=2cm, text width=2cm, align=center] {Textbook\\Selection};
\draw [arrow] (dec1) -- node[anchor=north] {$n$ = 4} (dec2);
\node (pro1) [process, below of=dec2, yshift=-1cm, text width=1.4cm, align=center] {Corpus Cleaning};
\draw [arrow] (dec2) -- node[anchor=west] {$n$ = 1} (pro1);
\node (pro2) [process, below of=pro1, yshift=0cm, text width=2cm, align=center] {Corpus Annotation};
\draw [arrow] (pro1) -- (pro2);
\node (pro3) [process, left of=pro1, xshift=-4cm, text width=2cm, align=center] {Collecting PGN};
\draw [arrow] (pro1) -- node[anchor=south] {$g$ = 91} (pro3);
\node (pro4) [process, left of=pro1, yshift=-2cm, xshift=-2cm, text width=2cm, align=center] {Creating FEN};
\draw [arrow] (pro1) -| (pro4);
\node (pro5) [process, below of=pro2, yshift=0cm, text width=2cm, align=center] {Topic Relevance};
\draw [arrow] (pro2) -- node[anchor=west] {$s$ = 1164} (pro5);
\node (pro6) [process, left of=pro5, xshift=-3cm, text width=2cm, align=center] {Sentiment Analysis};
\draw [arrow] (pro5) -- node[anchor=south] {$r$ = 782} (pro6);
\end{tikzpicture}}
\caption{The workflow followed in developing our corpus: data collection, processing and analysis. Key: $n$ = number of books, $g$ = number of games, $s$ = number of sentences, $r$ = number of topic relevance sentences.}
\label{figure:3}
\end{figure*}

Figure \ref{figure:3} summarises the steps taken in constructing the corpus. We searched Project Gutenberg (\url{https://www.gutenberg.org}), a free electronic textbooks library, using the search term ``chess'' and 25 e-books were retrieved and manually scanned according to the following selection criteria: 
\begin{itemize}
\itemsep0em
\item E-book must be in the English language.
\item E-book must be aimed at teaching how to play strategic chess moves. 
\item E-book must not be about the history of chess.
\item E-book must contain textual descriptions of moves, and not only listing moves from played games.
\item E-book must be aimed at teaching humans, and not for designing chess systems.
\item The author must be a chess master-player with an Elo rating above 2400. 
\end{itemize}
ELO is a rating system that relatively measures the skill levels of chess players. 2400 is a rating of most international master players and some Grandmaster players. Therefore, we can obtain a level of knowledge close to the rating of a strong chess engine, such as Stockfish (\url{https://stockfishchess.org/}), which can reach level up to superhuman with ELO rating above 3000 \cite{silver2017mastering}. The textbooks that met the initial search criteria are listed in Table  \ref{table:1}. 
After manual checking, we selected "Chess Strategy" textbook, E-book id (5614), by Edward Lasker, an international master chess player with an ELO rating of 2489. The textbook explains strategies played in popular tournaments games, and discusses moves that should, would, or not, have been considered. Also, this textbook speaks to both beginners and advanced players. 

Language grounding is required for a model to learn the representation of the knowledge in both the textual descriptions and the environment. We used regular expression-based rules to parse diagrams of the board states and convert them into the Forsyth–Edwards Notation (FEN) format. This is a chess notation that describes any board state of a game with pieces positions, player's turn and move. Chess engines use this format to initialise the game at any state. Finally, we retrieved the tournament games described in the textbook from the chess database (\url{https://www.chessgames.com/index.html}) in Portable Game Notation (PGN), a format that records moves, players' names, time/date and game result. Also, we manually created PGN files for games that were not retrievable from the database or from other sources.\



\subsection*{Data cleaning}
Descriptive notation is a move-recording style that was used until 1980. Since then, a new notation style, Standard Algebraic Notation (SAN), was introduced for machines to parse chess moves and games. To follow standardised chess semantics, we applied cleaning and preprocessing steps to convert descriptive notation to its corresponding SAN:
\begin{itemize}
\itemsep0em
\item Renaming of positions of pieces, such as columns names, e.g., from ``QR'' file to ``a'' file.
\item Changing descriptive notation of piece names, such as ``QR'' to ``Rook''.
\item Changing descriptive notation of movements to standard algebraic notation, such as ``QR2'' to ``Ra2'', for chess engine readability purposes.
\item Manual correction of incorrect mentions of a move or a piece in board diagrams, arising from optical character recognition (OCR) conversion errors.
\item Removal of diagrams and text sections that are not pertinent to a particular game, e.g., diagrams that were included in the textbook to illustrate piece movements without the other pieces presented in the board.
\end{itemize}

\subsection*{Data processing}

\subsubsection*{Annotation process for evaluating chess move} \label{sec:annotation}

Textbooks contain different types of sentences, such as introductory sentences which are not relevant to our task. Thus, we followed a similar approach for sentiment annotation as in \cite{toprak2010sentence, maehlum2019annotating} to annotate the corpus with move relevancy labels on a sentence level.  
A sentence is considered relevant only if it discusses a move, or a sequence of moves, as a topic of the sentence in an evaluative form. An example of a relevant sentence is ``\textit{To convert it into a win by queening the extra pawn is only a matter of time.}'' because it discusses the move ``\textit{queening the extra pawn}'' as a positive move because it will lead to "\textit{winning}". An example of non-relevance sentence is "\textit{We have now seen how the possession of open files reacts on the mobility of the opposing forces, forever increasing their difficulties until the positional advantage is converted into material gain.}".

Afterwards, each sentence labelled as relevant is annotated with one sentiment label with the aim to evaluate the move it discusses. For this task we followed the ``simple sentiment annotation schema'' described in \cite{mohammad-2016-practical}, applying it at the level of sentences. We define the sentiment labels as follows:
\begin{itemize}
\itemsep0em
    \item \textit{Positive}[label:2]: expresses a good outcome of playing a move for the player. An example sentence, "But White can, by a simple sacrifice, bring the slumbering R at a1 into sudden action: 1. ... Nxe4 2. Re1 Bf5 3. Nc3 Nd6 4. Rxe4 Nxe4 5. Re1 and White wins two pieces for his Rook."
    \item \textit{Negative}[label:0]: expresses a negative outcome of playing a move for the player. An example sentence, "An example of this is found in Diagram 6; Nxe4 fails on account of Rxc6; this leaves the Knight unprotected, and White wins two pieces for his Rook."
    \item \textit{Neutral}[label:1]: a sentence does not express any explicit outcome of a move. An example sentence, "It is Black's move, and we will suppose he wishes to play e5."
    \item \textit{Uncertain (not sure)}[label:3]: when the outcome of a move is difficult to identify, or it is difficult to identify an explicit move. An example sentence, "In both cases White has an easy development, whilst Black has no convenient square for his Queen's Bishop."
\end{itemize}

We consider negation in a sentence if it has a direct effect on its sentiment. For example, ``\textit{Black cannot very well exchange the pawns, leaving the King's file quite exposed, and must submit to White playing cxd5 maintaining the pawn at e4 and preventing Black's d5 for some time to come.}''. The negative polarity here implies negative sentiment toward the move ``\textit{exchange the pawns}''. The dataset was fully labelled by the first author, and a second annotator with chess domain background was recruited for measuring inter-annotator agreement.

\subsection*{Synthetic data generation} \label{sec:augmentation}
The size of the LEAP corpus is considered small compared to corpus sizes in other domains, and manually labelling more datasets is labor-intensive. Alternatively, data augmentation methods offer an option for increasing the corpus size, enabling language models to generalise and comprehend the contextual and linguistic characteristics of a specific domain. This, in turn, enhances the language model's performance in classification tasks. In this study, we adopted the data augmentation approach reported in similar research to generate synthetic data from the LEAP corpus \cite{li2023you}. We employed the DINO method introduced recently for this purpose\cite{dino} . Generative Pre-trained Transformer (GPT-2-xl) \cite{Radford2019LanguageMA} is the core component of the method, which follows unsupervised approach to generate synthetic data from scratch based on three prompt instructions: 
\begin{itemize}
    \item Write two sentences that mean the same thing.
    \item Write two sentences that are somewhat similar.
    \item Write two sentences that are on completely different topics.
\end{itemize}
The three prompt instructions are acted as a self-debiasing approach, where each prompt group would produce a sentence with a different meaning, and each generated sentence should be falling only into one prompt group to control the quality of the generated sentences. However, we acknowledge that the text generation model's output may differ from the original sentences, potentially resulting in the creation of illegal moves, altering the likelihood of making certain moves, or shifting the sentiment associated with those moves.

We generated 30 synthetic data (candidates) per original sentence (reference) per prompt instruction, resulting in a total of 99,529 generated synthetic data points. After removing duplicate sentences where the reference and candidate are the same, and eliminating sentences generated by the third prompt instruction based on irrelevant grounds, the final count of synthetic data was reduced to 82,145. 

To evaluate the quality of the synthetic data, we employed the BertScore metric \cite{BERTScore}. This metric represents each token using contextual embeddings and measures the cosine similarity between the reference and candidate sentences. The results were reported in terms of F1-score. Additionally, we utilised the BLEURT score \cite{bleurt}, a slightly modified version of the original Bert model, where the model was fine-tuned for language robustness and domain generalisation through an unsupervised approach, using millions of synthetic data pairs (reference, candidate). These pairs were generated using three techniques: different methods of mask filling, back translation from English to another language and back to English, and random word dropping from the generated data (reference sentence). The second modification involved leveraging the special token [CLS], which represents a vector of the contextual representation of the reference and candidate sentences. A linear layer was added on top of this representation to predict the similarity rating.

\begin{figure*}[h]
\centering
   \includegraphics[width=11cm, height=8cm]{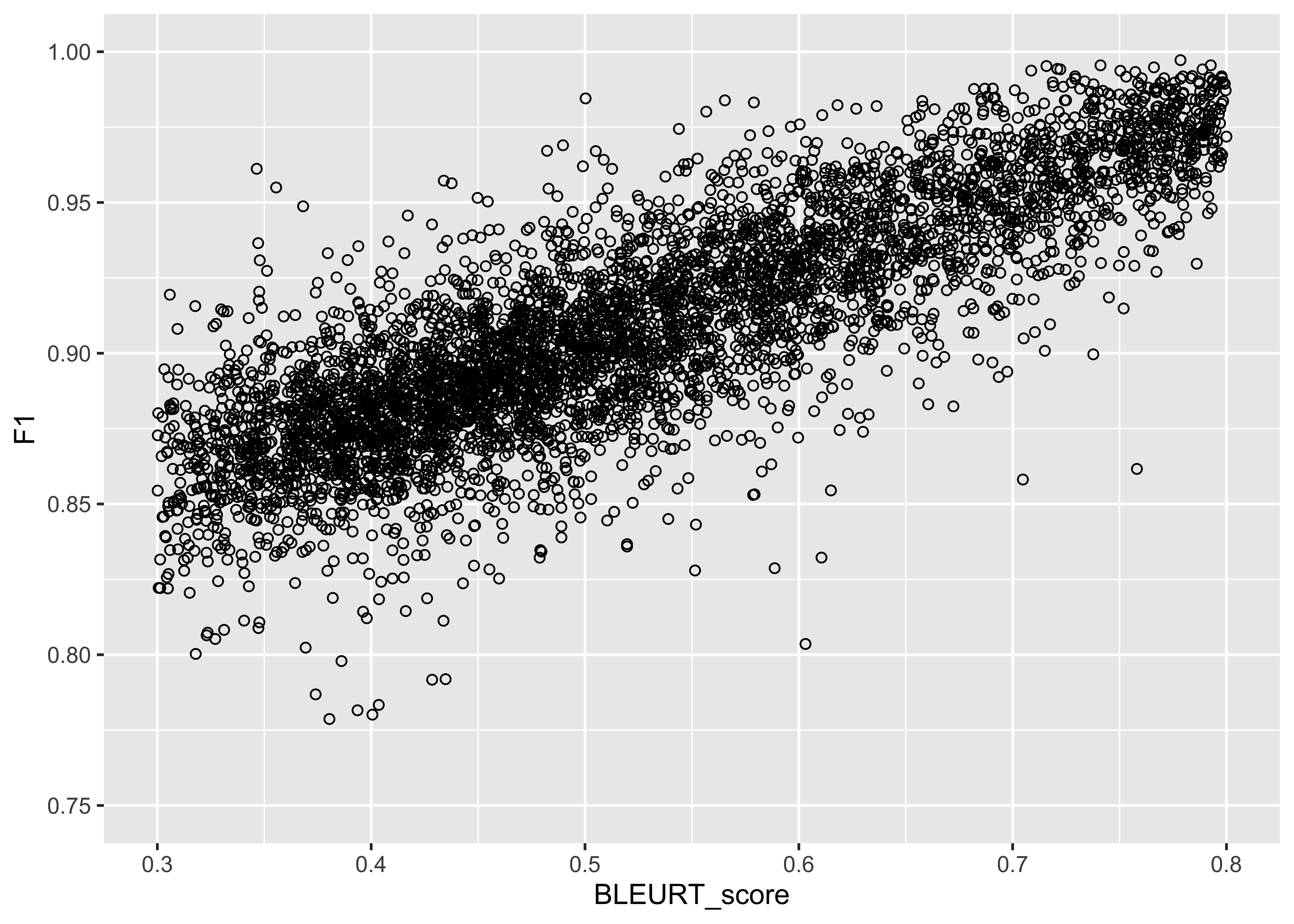}
   \caption{A random subset of 5000 synthetic sentences that were evaluated using BertScore and BLEURT metrics to demonstrate the correlation of the level of increasing between both scores.}
   \label{figure:4}
\end{figure*}

During evaluation and manual inspection of a subset of sentence pairs (reference, candidate), we observed that if the BLEURT score of the candidate sentence was above 80, it mostly indicated a duplicate of the reference sentence with limited changes, such as altering a number or removing a comma. To prevent the model from overfitting on duplicate sentences during training, we set the threshold for synthetic data with BLEURT scores between 30 and 80. This resulted in 69,049 synthetic sentences for the relevant classification task and 39,449 synthetic sentences for the sentiment classification task. The threshold range was selected after multiple experiments with different ranges and manual inspection of the data. Furthermore, we observed a relatively similar scoring range between both metrics, where high or low scores were assigned to instances by both metrics. To illustrate this observation, we visualised the scores from both metrics for a sample of 5,000 synthetic sentences in Figure \ref{figure:4}.

\subsection*{Classification models}
Recently, transformer architectures, based on the attention mechanism \cite{vaswani2017attention}, have achieved state-of-the-art (SOTA) results and outperforming any previous text classification models. Such models were pre-trained on large size of general domain datasets to gain large knowledge and transfer it into specific tasks. One advantage of adapting transfer learning technique is pushing SOTA results for specific task with labelled data in limited-resources domains, where there are not enough data to pre-train a model from scratch. Thus we selected four transformer-based pre-trained models based on various architectures, as baselines for both topic relevance and sentiment analysis classification tasks: (1) BERT \cite{devlin2018bert} was the first deep Bi-directional Encoder Representation using Transformer model developed based on masked language model (MLM) technique, which randomly masks words and trains to predict them based on the context embeddings. Also, it introduces Next Sentence Prediction (NSP) approach to improve the performance for natural language understanding tasks, such as Natural Language Inference (NLI). BERT obtained new state-of-the-art results on 11 different tasks, including General Language Understanding Evaluation (GLUE) \cite{wang2018glue}, 
(SQuAD 1.1 \cite{rajpurkar2016squad}, SQuAD 2.0) 
 and Situations With Adversarial Generations (SWAG) \cite{zellers2018swag}.
(2) XLNET \cite{yang2019xlnet} is an auto-regressive transformer model that uses permutation language model approach of all embedding factorisation in order to overcome the defects of words dependency in MLM approach. It outperformed BERT models on reading comprehension tasks including question answering, text classification datasets and GLUE tasks. (3) RoBERTa \cite{liu2019roberta} a modified BERT architecture that pre-trained BERT for a longer time, longer sentence and a larger corpus. The architecture achieved higher performance on some GLUE tasks comparing to both BERT and XLNET. (4) ALBERT \cite{Lan2020ALBERT} is a lighter BERT version that uses two parameter-decreasing techniques; factorized embedding parameterization and cross-layer parameter sharing. The techniques are designed to reduce the problem of large model size, which leads to memory limitations and long time training. 
Also, it uses self-supervised loss technique for sentence-order prediction (SOP) instead of NSP to improve inter-sentence coherence for multi-sentence encoding tasks. ALBERT outperformed BERT over GLUE, SQuAD datasets and RACE benchmark for reading comprehension task \cite{lai2017race} with (1.4\% - 8.4\%) improvement. ALBERT also outperformed RoBERTa and XLNET on some of the GLUE tasks, and on SQuAD and RACE benchmarks. Finally, we included the distilled version of both RoBERTa and BERT models, which is a reduced-size model of the original BERT and RoBERTa models but faster and on a comparable level of performance with the original models \cite{sanh2019distilbert}. 

Each model was developed with different settings, including the size and type of the training dataset (e.g., books, news articles), the length of embeddings; whether the type of tokens used (cased or uncased) and the number of parameters, which resulted in various model sizes. To thoroughly examine the impact of these settings, we explored both model sizes and token types for each architecture. Specifically, we considered (BERT-base-uncased, BERT-base-cased, BERT-large-uncased, BERT-large-cased, BERT-large-uncased-whole-word, BERT-large-cased-whole-word, distil-BERT-cased, distil-BERT-uncased) for BERT, (XLNET-base-cased and XLNET-large-cased) for XLNET, (RoBERTa-base, RoBERTa-large, distil-RoBERTa-based) for RoBERTa, and ALBERT-base, ALBERT-large for ALBERT. 
Finally, we demonstrate the Transformer model's proficiency in comprehending chess context and highlight the effectiveness of transfer learning. Thus, we employed several linear machine learning baseline models; Random Forest (RF), Support Vector Machines (SVM), and Multi-layer Perceptron Neural Network (MLP). For each model, we utilised three types of pre-trained embeddings to represent the corpus semantically, hence have a fair comparison with Transformers models: The pre-trained embeddings are: pre-trained GLOVE embeddings \cite{pennington2014glove}, BERT embeddings, and Sentence-BERT (all-MiniLM-L6) embeddings \cite{reimers2019sentence}.


 

\section*{Data Records}

Table \ref{table:2} summarises the characteristics of our corpus: number of sentences, tokens, unique tokens, discussed board states, number of sentences labelled, and summarises the synthetic data size. With regard to organising the corpus, the raw text of the textbook was split into paragraphs at first. However, a paragraph is segmented every time to a sentence referring to a board state diagram is encountered, to allow for separately storing the description together with the corresponding board state in FEN format.
Nonetheless, every full game is also provided in PGN format. All sentences annotated with both relevance and sentiment labels were saved separately and provided in JavaScript Object Notation (JSON), a data representation format, which contains the sentences and the classification label. \\
We split the dataset into training, validation and test subsets using the bootstrap sampling technique \cite{7497471}, to overcome the data imbalance issue, and set the seed to 50 for reproducibility. For relevance classification we divided the corpus into training, validation and test subsets following a 80-10-10\% split \cite{maehlum2019annotating}; for sentiment classification the proportion is 70-10-20\% \cite{rosenthal-etal-2015-semeval}. The resulting number of sentences in each subset is described in Table \ref{table:3}. 
We evaluated the suitability of the synthetic dataset for fine-tuning models by assessing the performance of the models on an unseen testing set, which consisted of the original sentences from the LEAP corpus. To ensure a proper evaluation, we split the synthetic sentences into a training set (70\% of the data) and a validation set (30\% of the data). Notably, we excluded the candidate sentences generated from the testing set sentences to prevent any data leakage during the evaluation process.

\section*{Technical Validation}


\subsection*{Annotation analysis}
To validate the labels described in Table \ref{table:2}, the dataset underwent a comprehensive annotation process. Initially, the first author fully labelled the dataset. Subsequently, the dataset was double-annotated by two annotators to measure inter-annotator agreement. This process was carried out in two rounds. The level of agreement was quantified using Cohen's Kappa coefficient \cite{cohen1960coefficient} ($\kappa$) for both annotation tasks and both rounds, and the results are presented in Table \ref{table:4}.

In the first round, 10\% of the dataset was used for inter-annotator agreement assessment. The results indicated a moderate to substantial agreement strength\cite{landis1977measurement}. This initial round of agreement evaluation helped identify and address any issues with the annotation guidelines, ensuring their robustness. Subsequently, in the second round, 30\% of the dataset was double-annotated, resulting in an "almost perfect" level of agreement for both annotation tasks \cite{landis1977measurement}.

The annotation schema adopted ensures the identification of domain-specific knowledge---in the form of sentences discussing and evaluating chess moves and strategies---which helps in preventing noisy sentences from being propagated onto the move decision-making process. However, as with many other domains, there is some data imbalance in our corpus, which can be seen in number of topic relevance and non-relevance classes, and between the sentiment labels in LEAP corpus. Moreover, textbook sentences have a tendency to describe moves and strategies that will lead to positive outcomes. Thus, there are almost twice as many positive sentences compared to negative ones. Finally, we compiled the most common comments provided by the annotators during the manual annotation process. These comments align with the challenges identified in the previous section and may have an impact on the results of automatic classification. The comments are as follows:

\begin{itemize}
\item The sentence discusses moves for both players.
\item Difficulty arises in assigning a sentiment when multiple moves are presented in a sentence.
\item Difficulty arises in selecting a sentiment for implicit moves in a sentence.
\item Difficulty arises in interpreting a sentence that discusses a move without access to the board state.
\item Sentences exist with contradictory sentiments regarding the same move at a specific board state.
\end{itemize}

These comments provide valuable insights into the complexities and nuances of the annotation process, highlighting the potential challenges that could affect the accuracy of automatic classification.


\subsection*{Empirical evaluation}
\begin{figure}[h]
\centering
\begin{subfigure}[b]{1\textwidth}
\centering
\includegraphics[width=13cm, height=10cm]{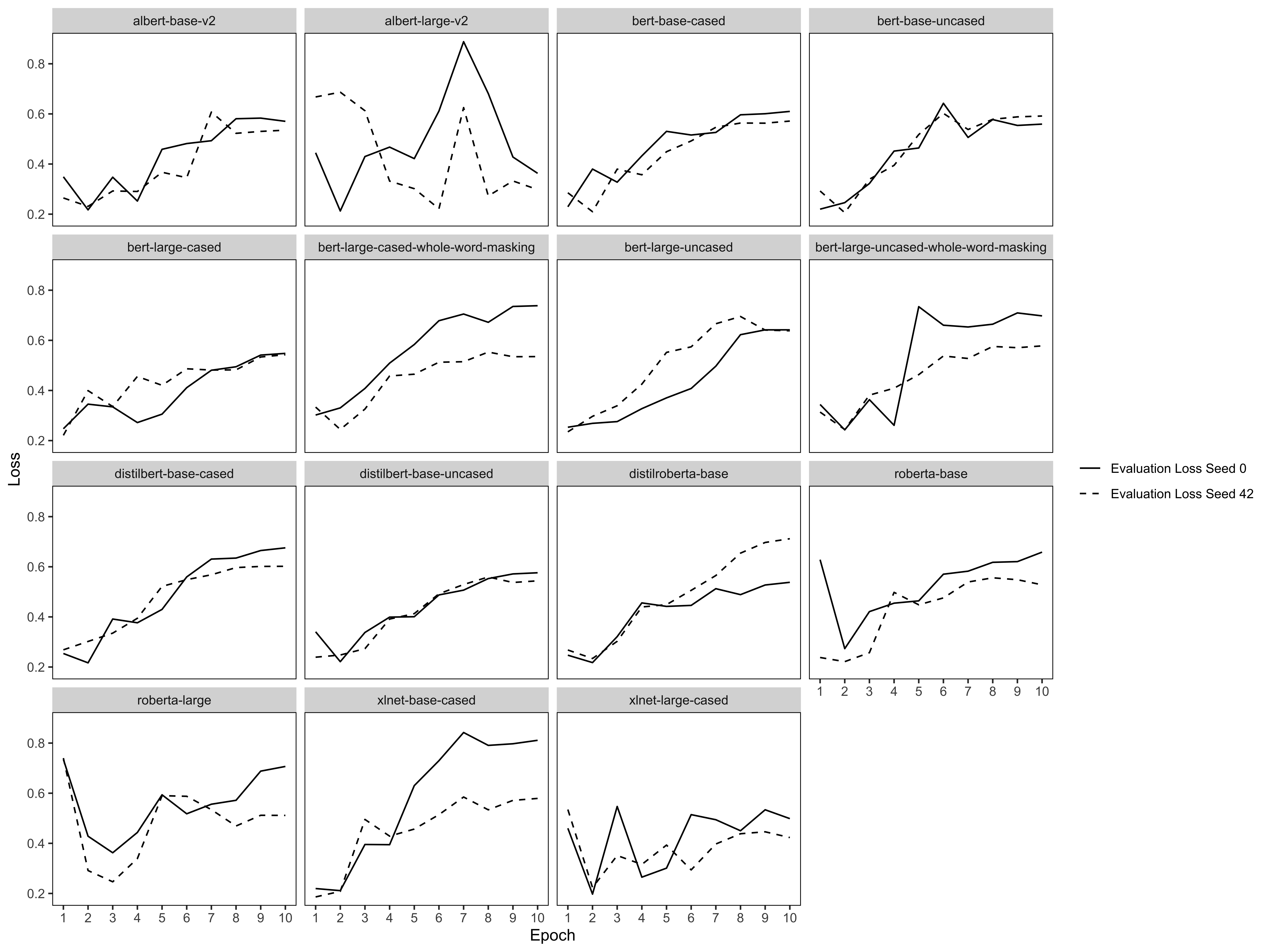}
\caption{Topic relevance classification task}
    \label{fig:5_a}
\end{subfigure}
\begin{subfigure}[b]{1\textwidth}
\centering
\includegraphics[width=13cm, height=10cm]{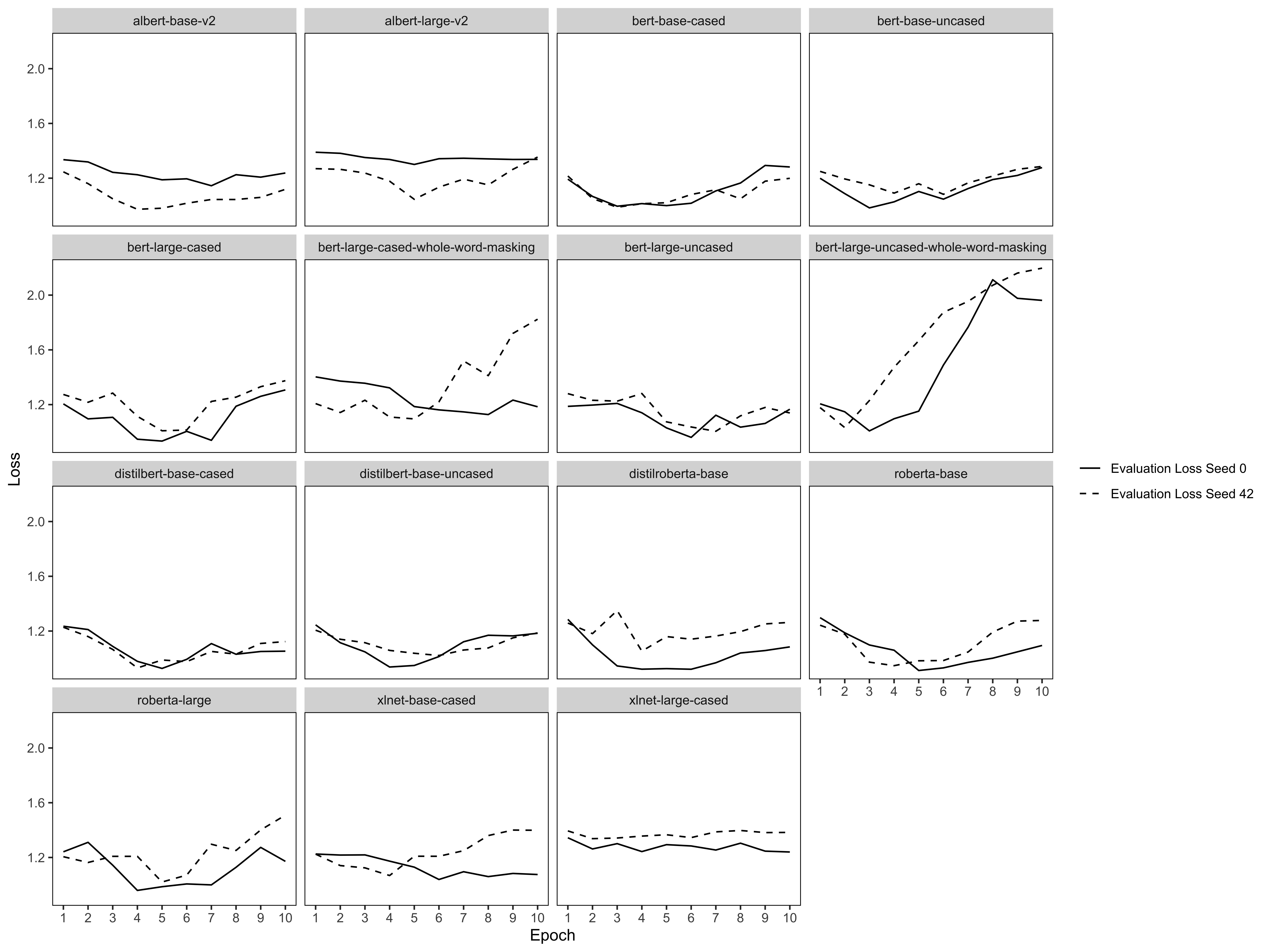}
\caption{Sentiment analysis classification task}
    \label{fig:5_b}
\end{subfigure}
\caption{Evaluation loss metric showing the learning curve of the transformer models for topic relevance classification over 10 epochs, which demonstrate the difference between number of best epoch per model with weight initialisation seed = 0 and seed = 42.}
\label{figure:5}
\end{figure}

\begin{figure}[h]
\centering
\begin{subfigure}[b]{0.45\textwidth}
\centering
\includegraphics[height=6cm, width=1\textwidth]{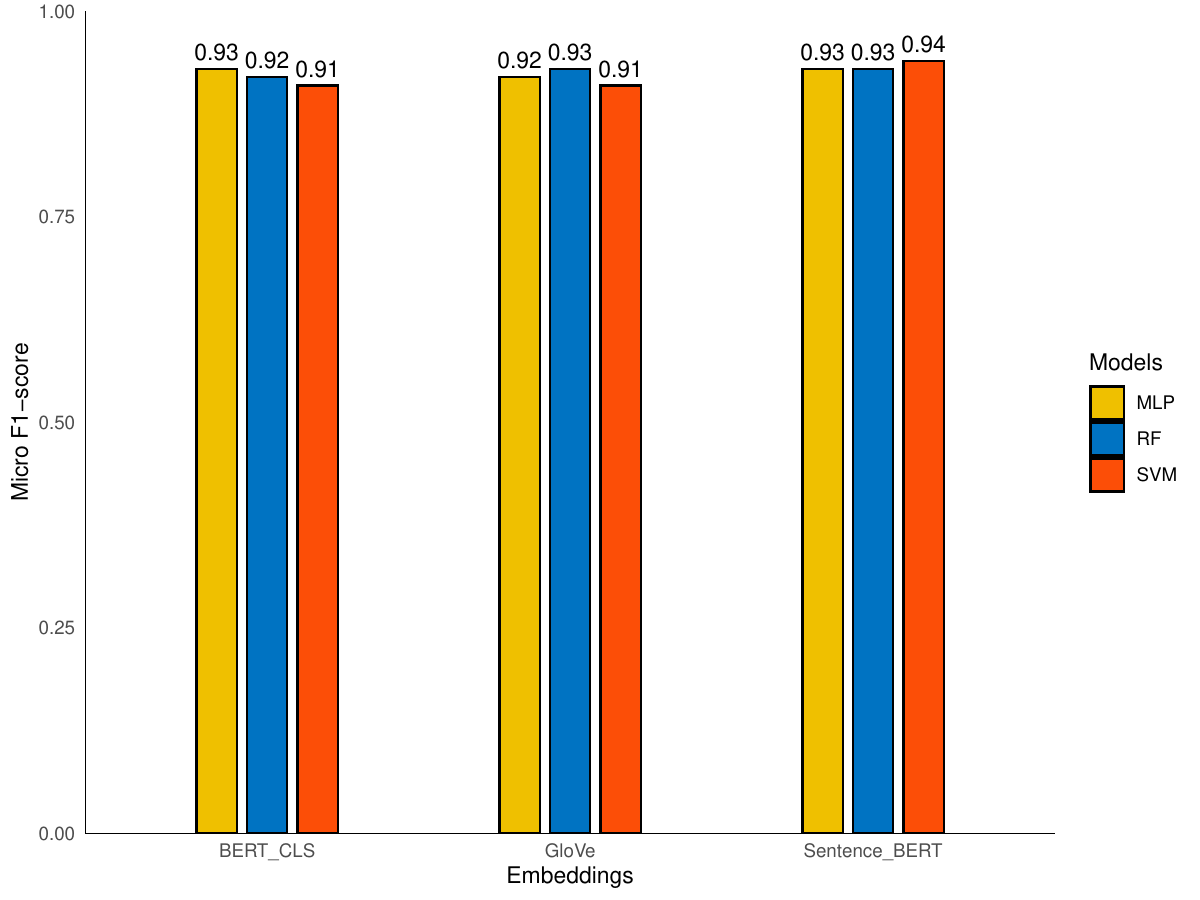}
\caption{}
\end{subfigure}
\begin{subfigure}[b]{0.45\textwidth}
\centering
\includegraphics[height=6cm, width=1\textwidth]{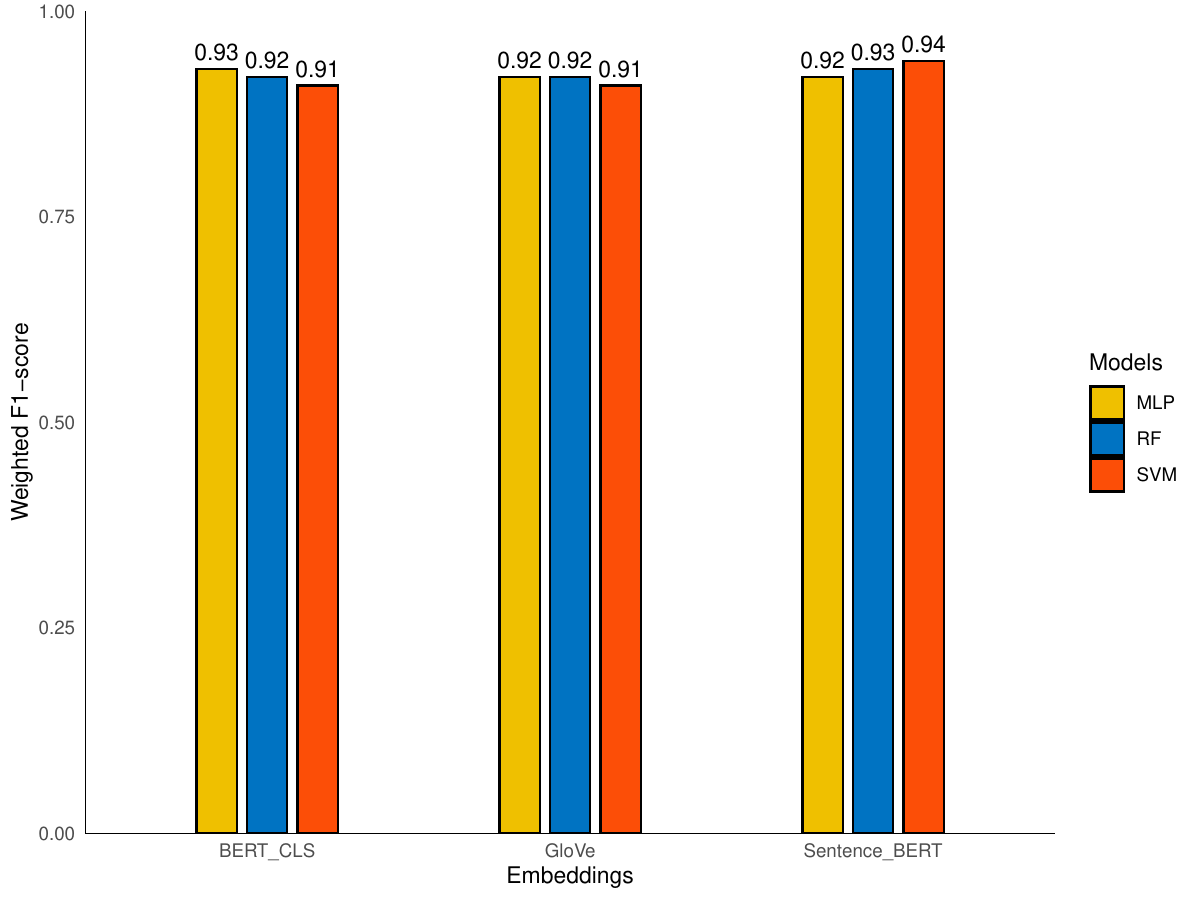}
\caption{}
\end{subfigure}
\begin{subfigure}[b]{0.45\textwidth}
\centering
\includegraphics[height=6cm, width=1\textwidth]{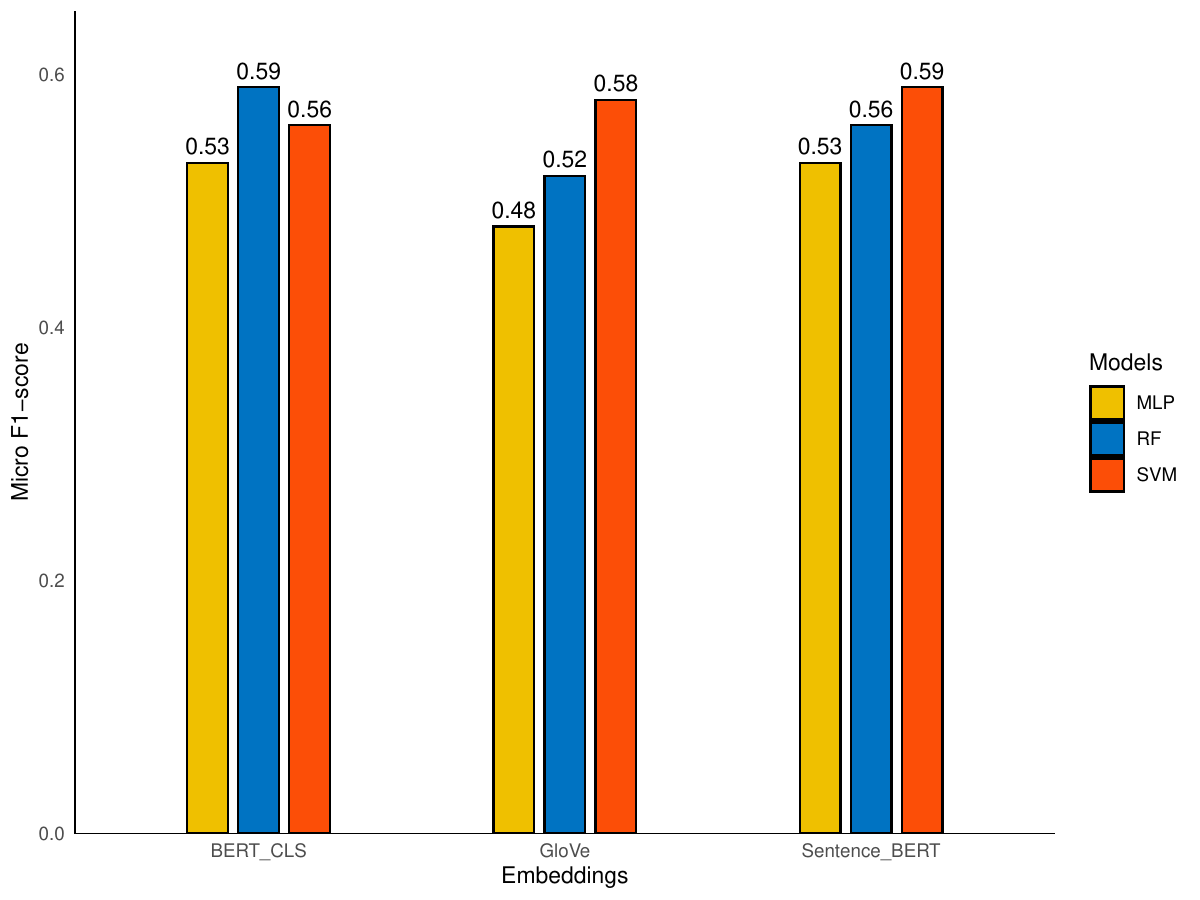}
\caption{}
\end{subfigure}
\begin{subfigure}[b]{0.45\textwidth}
\centering
\includegraphics[height=6cm, width=1\textwidth]{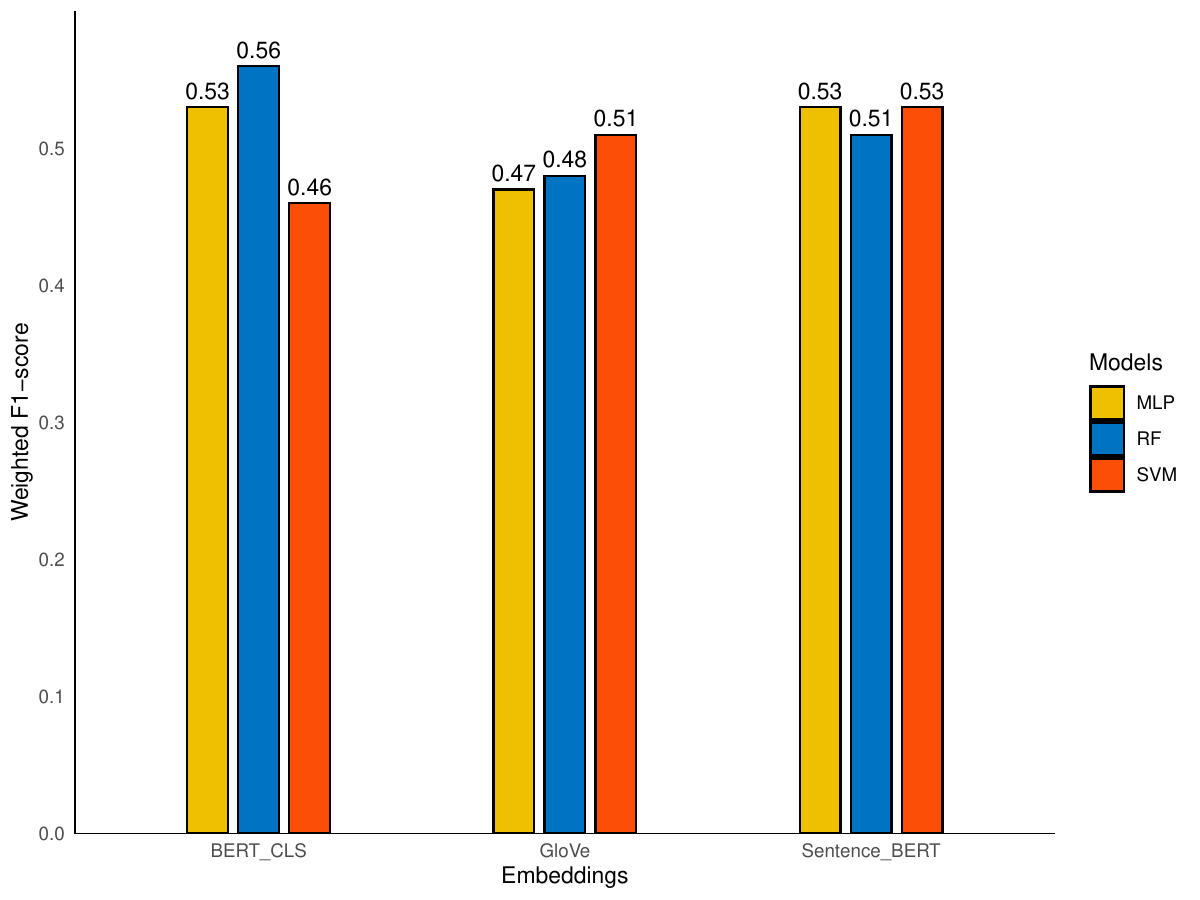}
\caption{}
\end{subfigure}
\caption{Micro and Weighted ${F}_{1}$ of three machine learning baseline models: Multi Layer Perceptron (MLP), Random Forest (RF), Support Vector Machine (SVM), each model was trained using Sentence-BERT embeddings, GLOVE embeddings and BERT base-uncased embeddings (CLS). Figures (a and b) for topic relevance classification, and Figures (c and d) for sentiment analysis classification.}
\label{figure:6}
\end{figure}

In this section, we describe the steps for evaluating our new Learning to Play Chess from Textbooks (LEAP) corpus, and the performance of state-of-the-art models in two classification tasks. The models hyper-parameters setting were the same for both classification tasks to understand the effect of the tasks and context on the models performance. We used the following baseline hyper-parameters for the Transformers models: learning rate: 4e-05, training and evaluation batch size: 8, dropout: 0.1. We randomly selected two weight initialisation seeds (0, 42) 
to understand if the ${F}_{1}$ scores are related to model sensitivity of randomisation for weight initialisation seeds or the scores are effected by classification task \cite{dodge2020fine}, and set the number of epochs to 10 for both classification tasks. During training phase, we evaluate the models at the end of each epoch using the validation set. The model "best epoch" is usually measured by achieving convergent; where the model achieves the lowest evaluation loss score when tested on the validation set and normally each model achieve the lowest evaluation loss at different epoch. Figure \ref{figure:5} summarises the evaluation loss functions over all epochs for both weights initialisation seeds and both classification tasks. The models obtained at the best epoch for both classification tasks were evaluated against the task corresponding testing set using weighted macro ${F}_{1}$ score, which take into account class imbalance in the dataset, and we included the micro ${F}_{1}$, which assign equal weight for all classes to measure the overall performance \cite{maehlum2019annotating, rosenthal-etal-2015-semeval}. The results of the Transformers models are summarised in Tables \ref{table:5} and \ref{table:7}, and the ${F}_{1}$ scores of the machine learning baseline models are reported in Figure \ref{figure:6}. 

Most Transformers models achieved between 88-97\% ${F}_{1}$ score on topic relevant classification task, and between 27-68\% ${F}_{1}$ score on sentiment classification task. Using BERT or Sentence-BERT embeddings slightly improved the performance of machine learning baseline models, however, almost all Transformer models achieved equal ${F}_{1}$ scores or a higher ${F}_{1}$ scores with 2-4\%. This clearly shows the power of the Transformer architecture and transfer learning approach in classification tasks with limited corpus size. Nonetheless, it is assumed that larger Transformer models usually lead to an improvement in the performance, regardless the size of the dataset used for fine-tuning \cite{devlin2018bert, Lan2020ALBERT}. Yet, such an improvements was not always present in both classification tasks, such as ALBERT large models achieved ${F}_{1}$ score less than baseline machine learning models in topic relevance task, and partially in sentiment analysis task, XLNet large model also achieved lowest ${F}_{1}$ score comparing to machine learning and Transformers models, and this observation was also presented in recent study \cite{guzman2022rafola}. One justification could be that the corpus size might affect the large models' performance. Also, different weight initialisation seed could negatively impact the model performance \cite{dodge2020fine} due to randomisation assignment of weights for deep learning models comparing to a more stable initialisation for machine learning models. Some large-based models were the most models sensitive to weight initialisation seeds, where ${F}_{1}$ score changed more than +\textbackslash- 10\%, e.g. ALBERT-large model and BERT-large-cased-whole-word-masking model in sentiment analysis task, and BERT-large-uncased for topic relevance task. However, such randomisation could be beneficial to achieve a higher performance, such as RoBERTa-base model achieving the highest ${F}_{1}$ score of 68\% using weight initialisation seed 42 comparing to 64\% ${F}_{1}$ score by DistilBERT-base-uncased using weight initialisation seed 0 for sentiment analysis classification task. We did not observe a direct affect of the token type (cased, uncased) to the ${F}_{1}$ scores, but they were mostly effected by the weight initialisation seed. This is an indication to not to follow the standard weight initialisation seed, which is usually 42, but to practice some experimentation with various models and careful selection of hyper-parameters, such as weight initialisation seed, before choosing.

We noticed that most models converged early in topic relevant classification task when epoch = 1 or 2, while the convergent in sentiment classification task was mostly on or after epoch 3. Also, the average evaluation loss function, on best epoch, achieved in topic relevant classification task was 0.24 for seed =0 and 0.22 for seed = 42. In sentiment classification task, the average evaluation loss function was 1.026 for seed = 0 and 1.041 for seed = 42. This shows that the models struggled in learning sentiment analysis task, which also explains the difference of ${F}_{1}$ scores achieved by the models between both classification tasks.

\begin{figure}[h]
\centering
\begin{subfigure}[b]{0.8\textwidth}
\centering
\includegraphics[width=0.77\textwidth]{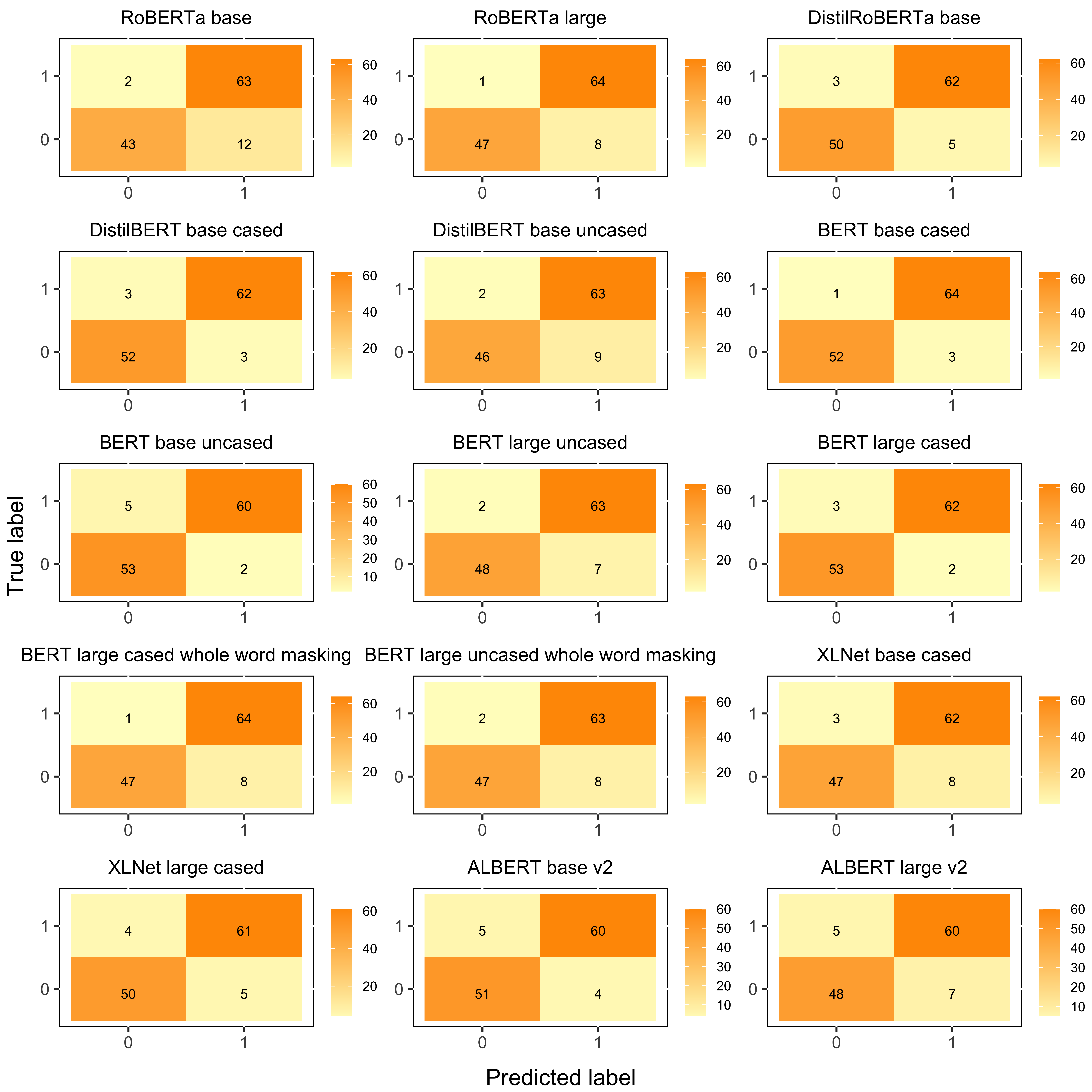}
\caption{}
\end{subfigure}
\begin{subfigure}[b]{0.8\textwidth}
\centering
\includegraphics[width=0.77\textwidth]{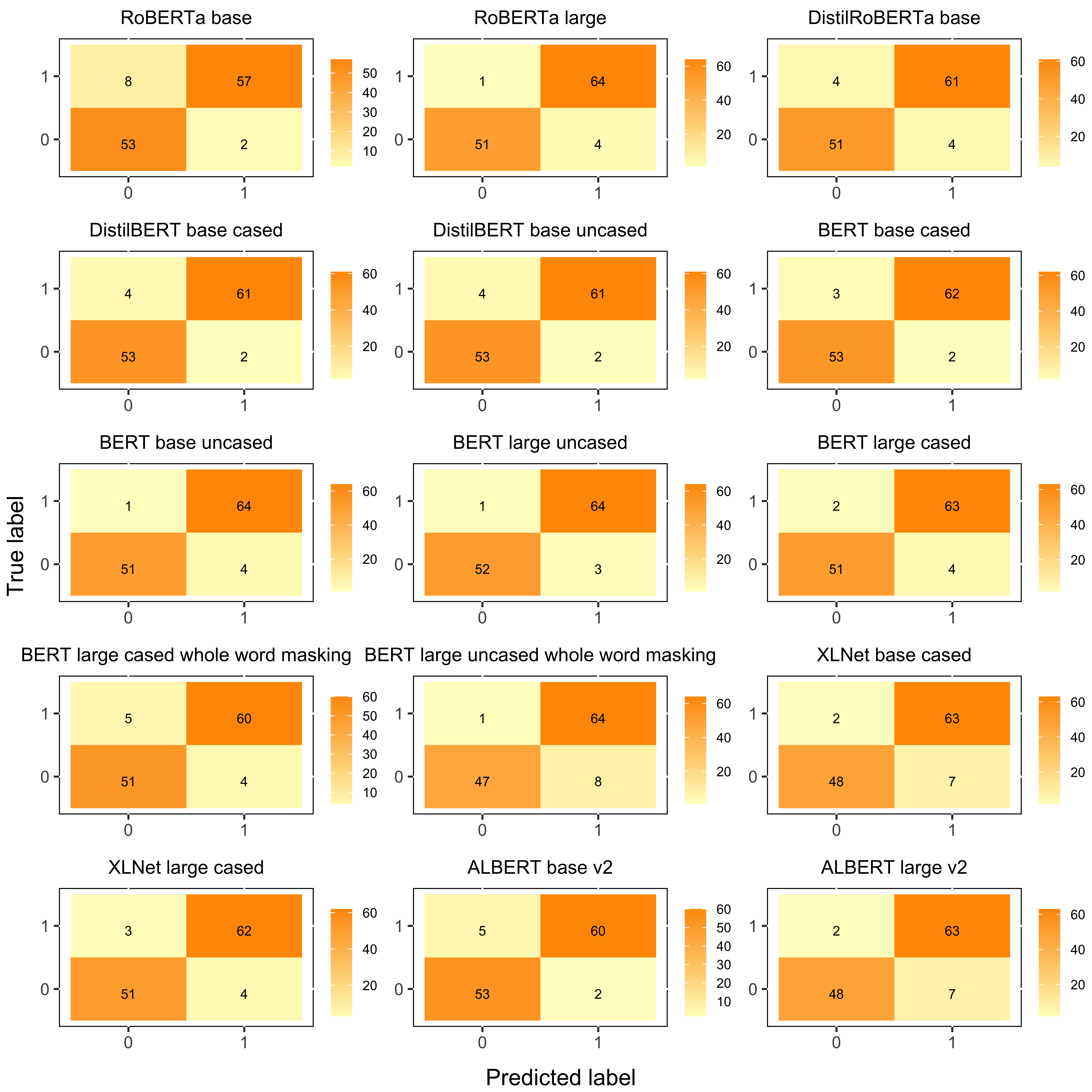}
\caption{}
\end{subfigure}
\caption{Results of models performance in label prediction on testing data comparing to true labels in topic relevant classification task. Models are in descending order by highest ${F}_{1}$ score with weight initialisation seed = 0 (a), and seed = 42 (b).}
\label{figure:7}
\end{figure}

\begin{figure}[h]
\centering
\begin{subfigure}[b]{0.8\textwidth}
\centering
\includegraphics[width=0.77\textwidth]{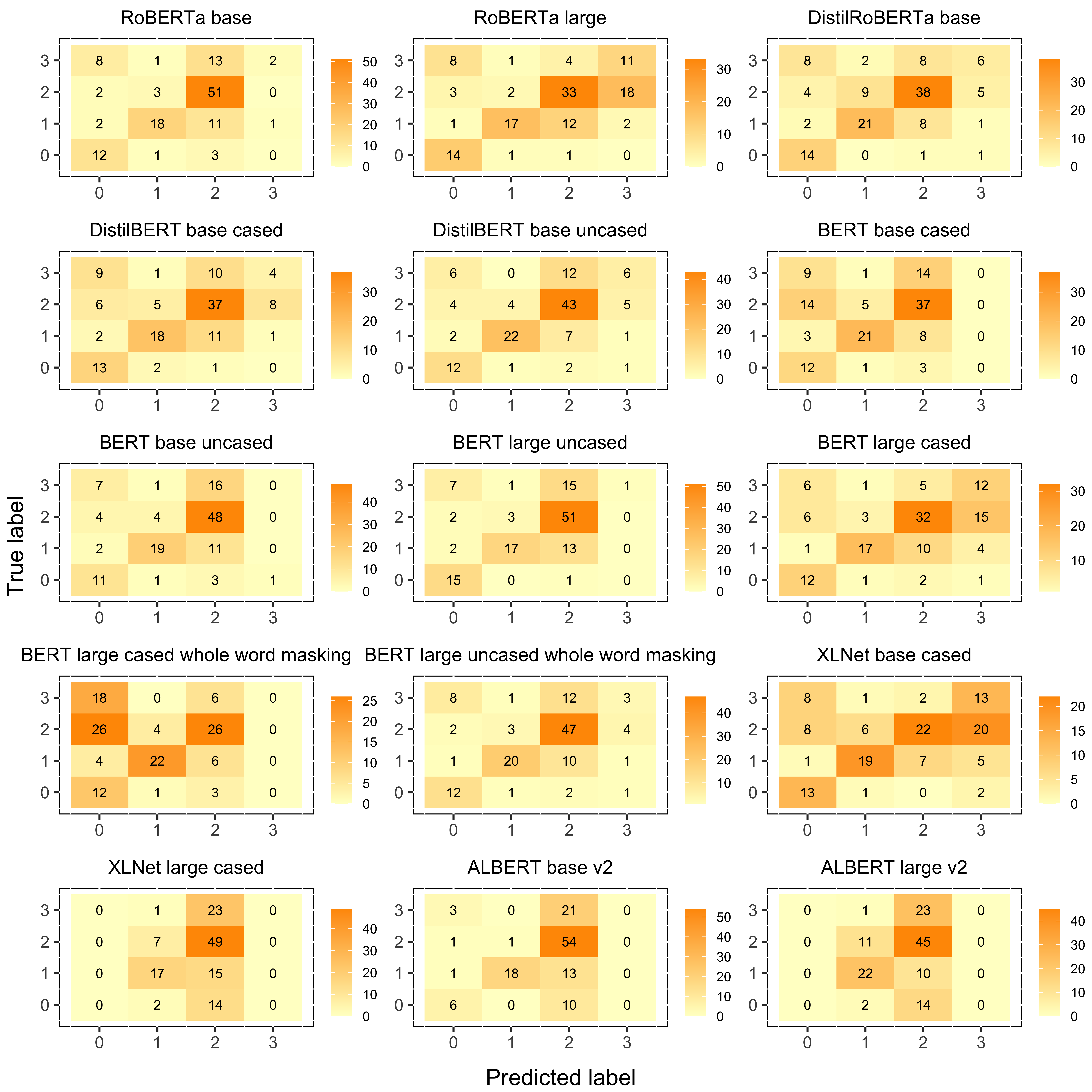}
\caption{}
\end{subfigure}
\begin{subfigure}[b]{0.8\textwidth}
\centering
\includegraphics[width=0.77\textwidth]{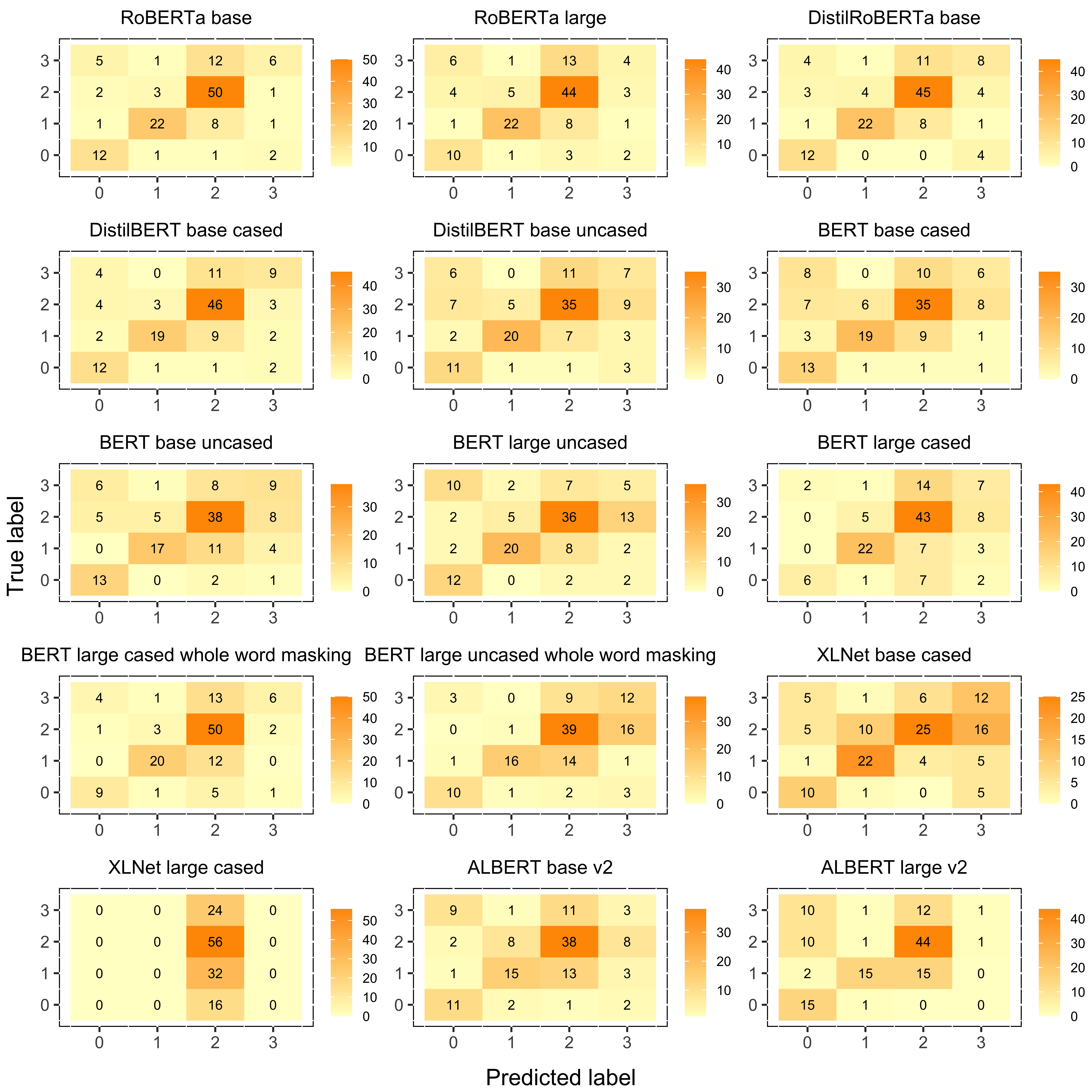}
\caption{}
\end{subfigure}
\caption{Results of models performance in label prediction on testing data comparing to true labels in sentiment analysis classification task. Models are in descending order by highest ${F}_{1}$ score with weight initialisation seed = 0 (a), and seed = 42 (b).}
\label{figure:8}
\end{figure}

To understand the effect of the context, first, we measured the reading comprehension of chess-teaching sentences by Flesch-Kincaid reading ease formula \cite{kincaid1975derivation}. The sentences' easy to read score was 93, with an average of 24 words per sentence, which indicates that humans can process and understand the text easily. Secondly, we analysed the models' ability to understood the context by label predictions distribution in both classification tasks. Also, we show a sample from the testing dataset of each task in Table \ref{table:9} and Table \ref{table:10}; 10 sentences for each correctly and incorrectly predicted class labels by the models that achieved highest f1-score per each seed. Figure \ref{figure:7} shows that models understood the context for labelling topic relevant classification task, while data imbalance issue might be a reason for false positive between both classes \cite{gao2020ensemble}. The same issue mostly contributed to lowering the ${F}_{1}$ scores in sentiment analysis classification task for all models, as shown in Figure \ref{figure:8}, due to small size of 'not sure [3]' class. In the same classification task, most false positive was between 'neutral [1]' and 'positive [2]' classes. Giving that chess is closed domain, many domain-specific terms are repeated, and sentence structure and semantic are relatively similar, regardless of the class label, which might confused the models. 

Also, many sentences discuss multiple moves or the effect of the move on both players, which sometimes confused the annotators and the models. On the other hand, sentences that discuss single move and its effect to a single player are more likely to be classified correctly, as seen in examples reported in Table \ref{table:10}. Hence, sentiment analysis schema on a finer-grained level, such as Aspect-based Sentiment Analysis (ABSA) \cite{xu2019bert} and semantic role labelling (SRL) level \cite{mohammad-2016-practical, yin-etal-2020-sentibert}, might improves the analysis by focusing on the move as a primary target of the sentiment. Finally, human understands the semantic of language by its environment, hence depending only on words for analysing moves, without access to the environment "board" that the text describes, hindered evaluating the quality of the moves in sentiment classification task \cite{beinborn-etal-2018-multimodal}. 

Finally, to address the effect of class imbalance and the corpus size on the Transformers models performance, we created random three sub-datasets using the synthetic data we generated from the LEAP corpus to fine-tuned the Transformer models; (1) "balanced": we used the original LEAP corpus with synthetic sentences to balance the number of classes, (2) "oversampled": original LEAP corpus over-sampled the minority classes with synthetic sentences to increase the size of the dataset and balance the classes as well, "synthetic": fine-tuned the models using only the synthetic dataset. Furthermore, to analyse if chess terminology have an impact over the models performance, we masked the chess entities of moves and players in the original LEAP corpus and replaced them with "MOV and PLY" terms. Each dataset was split into 70\% training and 30\% validation set, and we evaluated the fine-tuned models using the original LEAP testing set with the same hyper-parameters. 

\begin{figure}[h]
\centering
  \begin{subfigure}[b]{0.45\textwidth}
    \includegraphics[width=\textwidth]{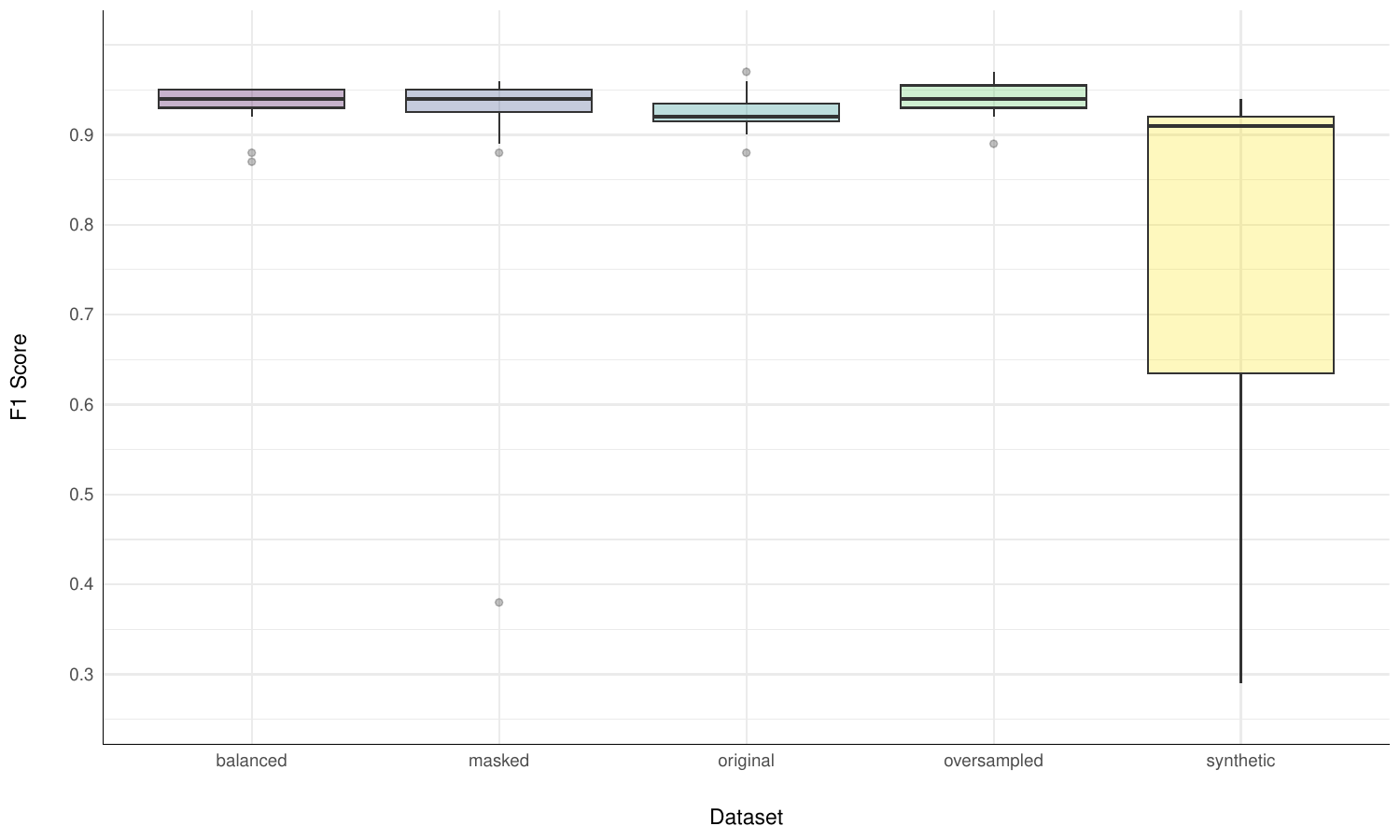}
\caption{Weight Initialisation Seed = 0}
    \label{fig:6_a}
\end{subfigure}
  \hfill
  \begin{subfigure}[b]{0.45\textwidth}
    \includegraphics[width=\textwidth]{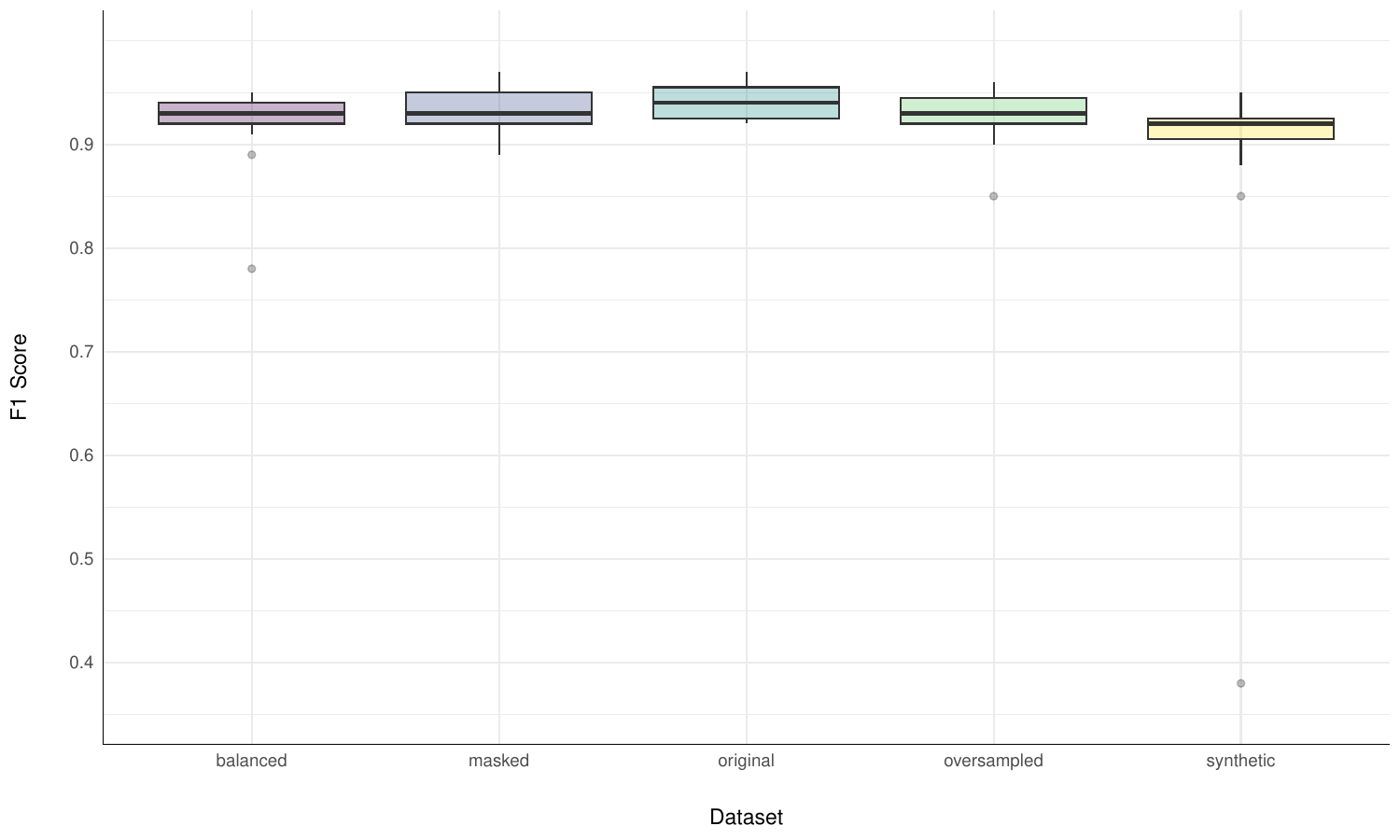}
    \caption{Weight Initialisation Seed = 42}
    \label{fig:6_b}
\end{subfigure}
    \vspace{0.8cm} 
  \begin{subfigure}[b]{0.45\textwidth}
    \includegraphics[width=\textwidth]{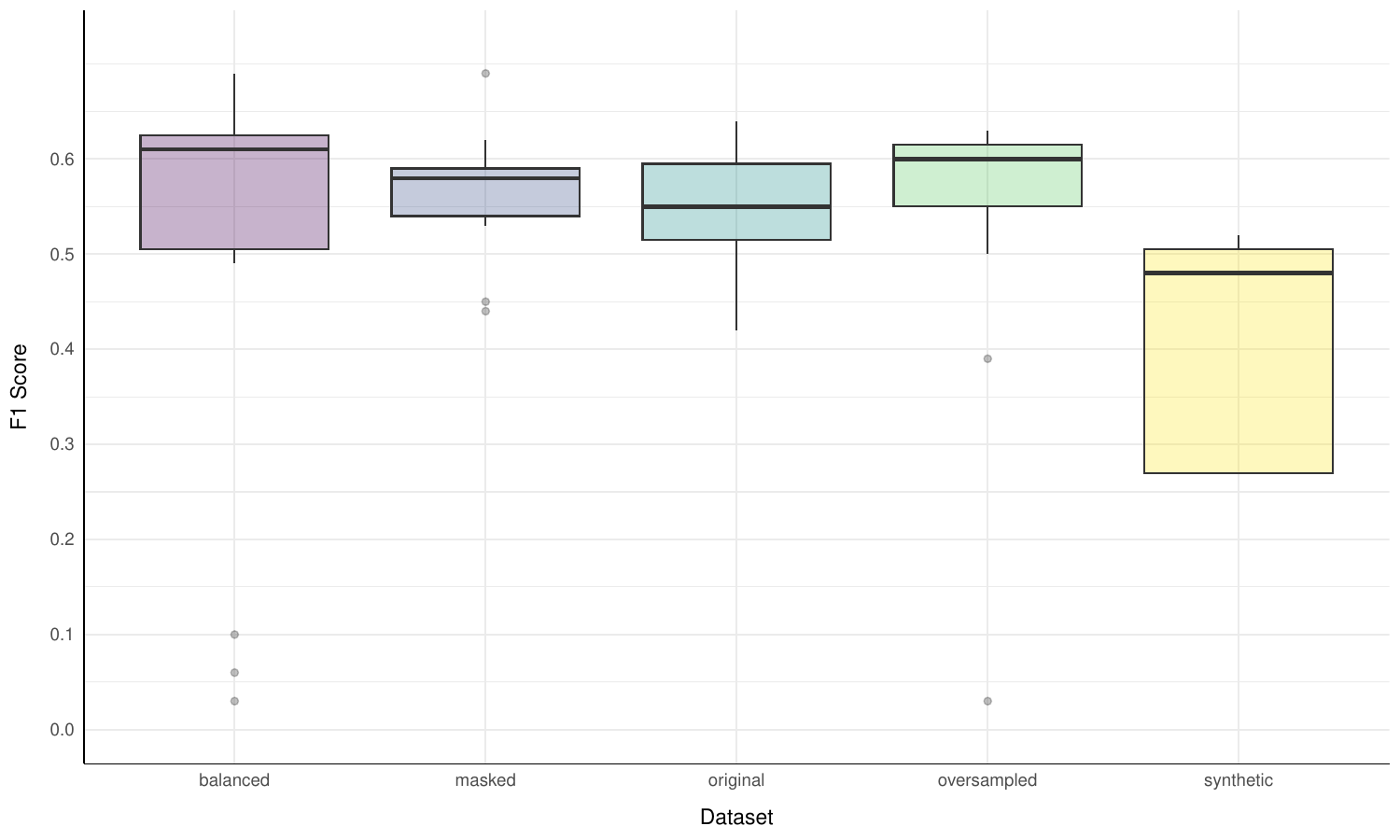}
    \caption{Weight Initialisation Seed = 0}
    \label{fig:6_c}
\end{subfigure}
  \hfill
  \begin{subfigure}[b]{0.45\textwidth}
    \includegraphics[width=\textwidth]{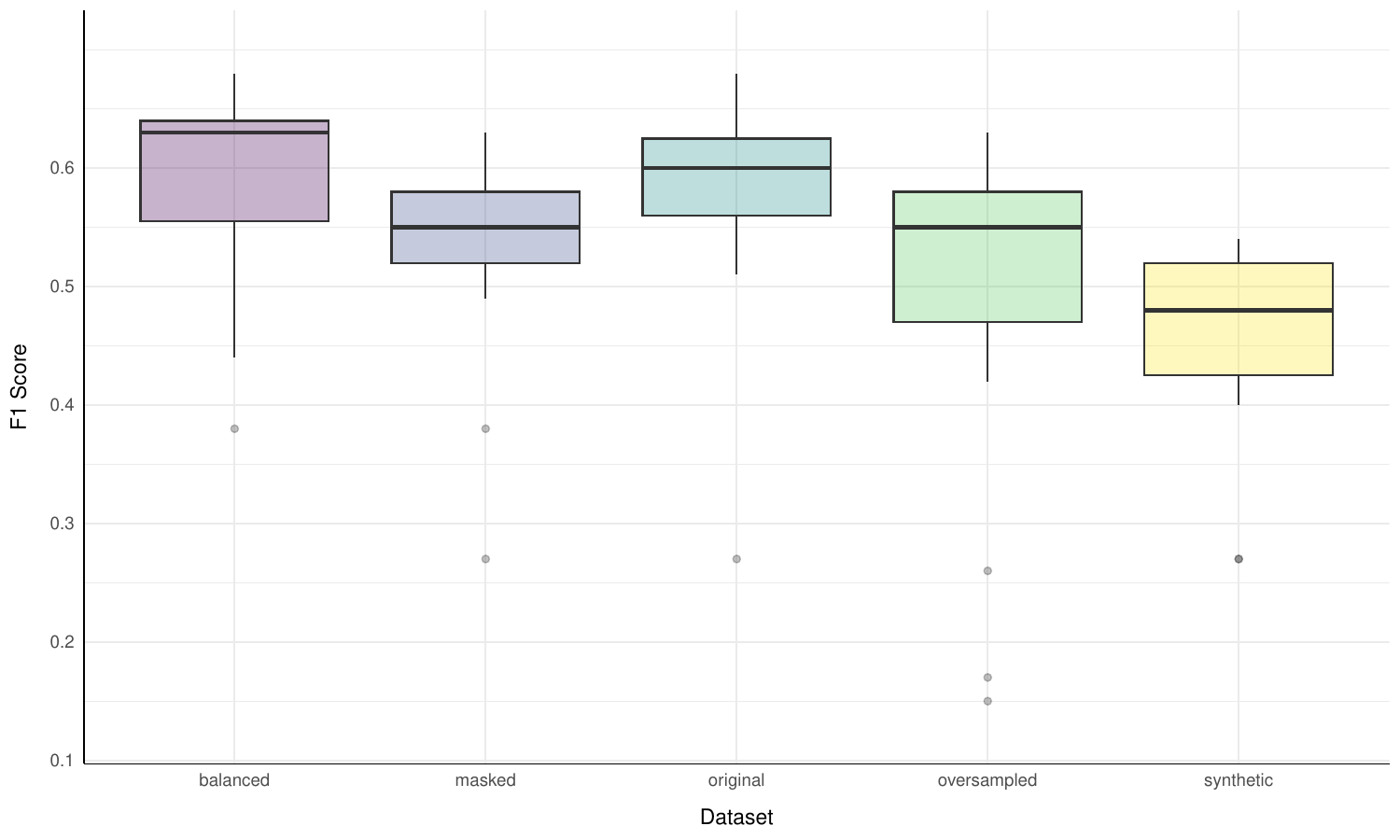}
    \caption{Weight Initialisation Seed = 42}
    \label{fig:6_d}
\end{subfigure}

\caption{Box plot showing the Weighted F1 scores of the Transformer models over the testing set after fine-tuning the models
with different settings using the training and validation sets; "balanced": original LEAP corpus with synthetic sentences to
balance the number of classes, "masked": original LEAP corpus with chess entities of moves and players are being masked and
replaced with "MOV and PLY" terms, "original": the original LEAP corpus, "oversampled": original LEAP corpus
over-sampled with synthetic sentences to increase the size of the dataset, "synthetic": the synthetic dataset only generated from
the LEAP corpus. Figures (a and b) for topic relevance classification task, Figures (c and d) for sentiment analysis classification
task.}

\label{figure:9}
\end{figure}

Figure \ref{figure:9} depicts the weighted macro ${F}{1}$ scores for both classification tasks using the five datasets. Firstly, the ${F}{1}$ scores of the Masked dataset indicate that the removal of chess entities does not necessarily affect or improve performance, especially in the sentiment analysis classification task. Secondly, as expected, balancing the classes and sometimes increasing the size of the dataset can enhance the ${F}{1}$ scores for both tasks. However, such improvement was limited, and the utilization of synthetic data to fine-tune the models resulted in a reduction of the ${F}{1}$ scores.

\begin{figure}[h]
\centering
   \includegraphics[width=11cm, height=8cm]{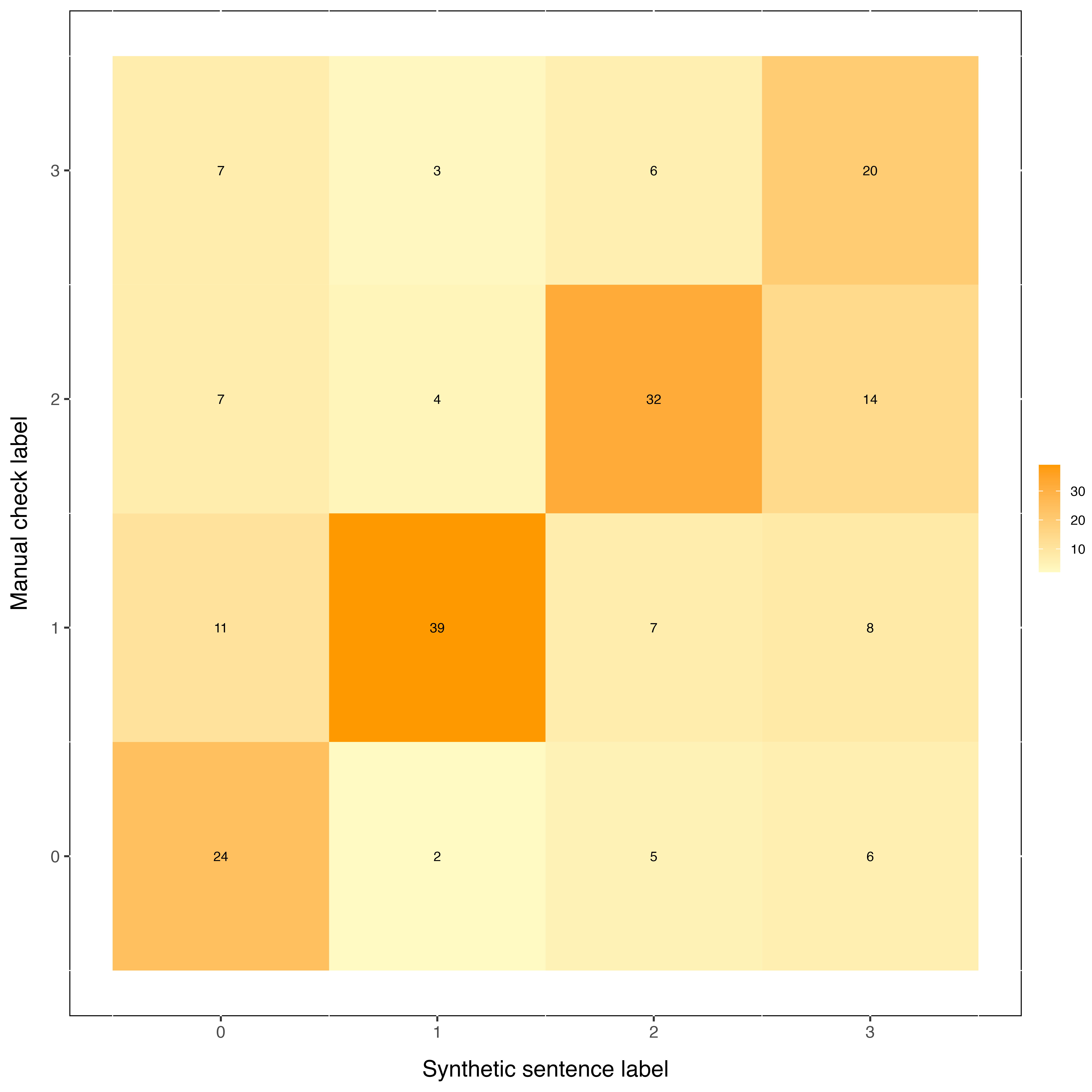}
   \caption{Confusion matrix of 200 synthetic sentence randomly selected between original sentence label and the manual label.}
   \label{figure:10}
\end{figure}

We manually analyzed and labeled 200 sentences randomly selected from the synthetic data for the sentiment analysis task. We found that it is not always possible to automatically transfer the original sentence label to the synthetically generated sentence. Although the generated sentence presents a high level of chess context, the underlying meaning of the original sentence that influences the classification label cannot always be the same for the synthetically generated sentence. We suspect that this confuses the models, and we illustrate this confusion using a confusion matrix between the manual labels and the original sentence labels in Figure \ref{figure:10}, along with a sample of the sentences in Table \ref{table:11}. It is evident that some sentences originally labeled as 'negative [0]' sentiment were changed to 'neutral [1]' sentiment in the synthetic sentence, and some original sentences labeled as 'not sure [3]' were transformed into 'positive [2]' sentiment in the generated sentence. Thus, the models are likely to be fine-tuned using incorrect labels that do not correspond to the context of the sentence, resulting in incorrect predictions during testing.

Furthermore, the Cohen's Kappa coefficient \cite{cohen1960coefficient} between the labels was ($\kappa$) = 0.45, with 4 synthetic sentences marked as not topic-relevant, and 115 sentences having matching labels. This indicates a low coefficient between the labels, and therefore, we cannot rely on automatically transferring the original sentence label to the synthetic one. However, synthetic data can be employed to enrich the models' understanding of chess context, as demonstrated by the ${F}_{1}$ scores obtained using the "balance" and "oversampled" datasets.

In conclusion, considering the inherent difficulty of the tasks, synthetic data did not significantly improve performance in the classification tasks. However, it proved valuable in enhancing chess knowledge and as a cost-effective alternative to processing chess textbooks in the traditional manner, offering advantages in terms of cost and Optical Character Recognition (OCR) correction.


\section*{Usage Notes}



\section*{Code availability}

All provided code were tested and run on a CPU with Python 3 (version 3.6 and above). The pre-trained models were fine-tuned using the Python-based the HuggingFace Transformers library \cite{wolf2020transformers} (version 4.10).  We used 4 Nvidia Volta v100 GPUs in fine-tuning the pre-trained models. For reproducibility, we provide the code for classification tasks, and all datasets (raw, split, synthetic), annotation guidelines and the models evaluation are free to use and available in the repository (\url{https://github.com/resrepos/LEAP}). 


\bibliography{main}


\section*{Acknowledgements}
We thank Ross King for discussing the idea of the project and for his early feedback in the project. We thank all the annotators who participated in the annotation tasks.


\section*{Author contributions statement}
H.A. conceived and conducted the experiments, analysed the results and drafted the manuscript. R.B. revised the analysis, reviewed the manuscript and supervised all steps of the work. All authors read and approved the final version of the manuscript. 

\section*{Competing interests} 
The authors declare no competing interests.

\section*{Figures \& Tables}






\begin{table}[h!]
\centering
\caption{Chess textbooks retrieved from Project Gutenberg library}
\label{table:1}
\begin{tabular}{p{1.5cm}p{2.5cm}p{6.5cm}p{3.5cm}}
\hline
E-book ID & Popularity & Book Name & Author \\\hline
33870 & 1026 downloads & Chess Fundamentals & Jose Raul Capablanca \\
5614 & 457 downloads & Chess Strategy & Edward Lasker\\
16377  & 223 downloads & The Blue Book of Chess & Howard Staunton \\
4913 & 82 downloads & Chess and Checkers: The Way to Mastership & Edward Lasker \\\hline
\end{tabular}
\end{table}

\begin{table}[h!]
  \centering
    \caption{Original curated dataset size after annotation and Synthetic Dataset Size. The data augmentation helped to reduce the effects of class imbalance issue, which provided (43\%) non-relevant data and (57\%) relevant sentence, in which (22\%) sentences labelled with negative sentiment, (19\%) sentences labelled with neutral sentiment, (38\%) sentences labelled with positive sentiment and (21\%) sentences labelled with not sure.}
    \label{table:2}
    \begin{tabular}{p{8.5cm}p{1.5cm}}
    \hline
    Dataset description & Size \\\hline
    Sentences & 1164 \\
    Tokens & 28139 \\
    Unique tokens & 2829 \\
    Games & 91 \\ \hdashline
    Original sentences discussing moves (topic-related) & 673 \\ 
    Original sentences not discussing moves (non topic-related) & 491 \\ \hdashline
    Original topic-related sentences with positive sentiment & 258 \\
    Original topic-related sentences with negative sentiment & 134 \\
    Original topic-related sentences with neutral sentiment & 150 \\ 
    Original topic-related sentences with not sure sentiment & 131 \\ 
    \hdashline
    Synthetic sentences discussing moves (topic-related) & 39449 \\ 
    Synthetic sentences not discussing moves (non topic-related) & 29600 \\ \hdashline
    Synthetic topic-related sentences with positive sentiment & 14835 \\
    Synthetic topic-related sentences with negative sentiment & 8700 \\
    Synthetic topic-related sentences with neutral sentiment & 7552 \\ 
    Synthetic topic-related sentences with not sure sentiment & 8362 \\ \hline
    \end{tabular}
\end{table}

\begin{table}[h!]
  \centering
    \caption{Topic relevance (move) classes and sentiment analysis labels for move evaluation per dataset splittings; training, validation and testing. \textbf{(O)} refers to the original datasets of the LEAP corpus and \textbf{(S)} refers to the synthetic dataset generated from the original training and validation sets only, excluding sentences from the original testing set.}
    \label{table:3}
\begin{tabular}{lllllll}

\toprule
 & \multicolumn{2}{c}{Topic relevance (move) classes} & \multicolumn{4}{c}{Sentiment analysis labels for move evaluation}\\
 \cmidrule(lr){2-3}\cmidrule(lr){4-7}
Dataset & relevant sentence & non relevant sentence & negative & neutral & positive & not sure \\
\midrule 
Training set \textbf{(O)} & 532 & 372 & 103 & 101 & 170 & 90\\
Training set \textbf{(S)} & 32030 & 23691 & 6704 & 5339 & 10482 & 6119\\

Validation set \textbf{(O)}& 76 & 64 & 15 & 17 & 32 & 17 \\
Validation set \textbf{(S)}& 3912 & 2976 & 842 & 630 & 1340 & 732 \\

Testing set \textbf{(O)}& 65 & 55 & 16 & 32 & 56 & 24 \\
\bottomrule
\end{tabular}
\end{table}

\begin{table}[h!]
  \centering
      \caption{Inter-annotator agreement for the two labelling tasks; topic relevance and sentiment analysis, using Cohen's Kappa coefficient.}
      
    \label{table:4}
    \begin{tabular}{p{4cm}p{3.5cm}p{3.5cm}}
    \hline
    Rounds & Topic Relevance $\kappa$ & Sentiment Analysis $\kappa$\\ \hline
    Round 1 (annotator 1 \& 2) & 50\% & 70\% \\
    Round 2 (annotator 1 \& 3) & 86\% & 87\% \\ \hline
    \end{tabular}

\end{table}

\begin{table}[h!]

  \centering

  \caption{Micro ${F}_{1}$ and Weighted Macro ${F}_{1}$ scores of transformer models for topic relevance classification task using original dataset. }

  \label{table:5}

  \begin{tabular}{lllll}

  \toprule
   & \multicolumn{2}{c}{Seed 0} & \multicolumn{2}{c}{Seed 42}\\

   \cmidrule(lr){2-3}\cmidrule(lr){4-5}

   Models & micro ${F}_{1}$ & weighted macro ${F}_{1}$ & micro ${F}_{1}$ & weighted macro ${F}_{1}$ \\

   \midrule
ALBERT base & 0.93 & 0.93 & 0.94 & 0.94 \\
ALBERT large & 0.90 & 0.90 & 0.93 & 0.92 \\
DistilBERT base cased & 0.95 & 0.95 & 0.95 & 0.95 \\
DistilBERT base uncased & 0.91 & 0.91 & 0.95 & 0.95 \\
DistilRoBERTa base & 0.93 & 0.93 & 0.93 & 0.93 \\
BERT base cased & 0.97 & \textbf{0.97} & 0.96 & 0.96 \\
BERT base uncased & 0.94 & 0.94 & 0.96 & 0.96 \\
BERT large cased & 0.96 & 0.96 & 0.95 & 0.95 \\
BERT large cased whole word masking & 0.93 & 0.92 & 0.93 & 0.93 \\
BERT large uncased & 0.93 & 0.92 & 0.97 & \textbf{0.97} \\
BERT large uncased whole word masking & 0.92 & 0.92 & 0.93 & 0.92 \\
RoBERTa base & 0.88 & 0.88 & 0.92 & 0.92 \\
RoBERTa large & 0.93 & 0.92 & 0.96 & 0.96 \\
XLNet base cased & 0.91 & 0.91 & 0.93 & 0.92 \\
XLNet large cased & 0.93 & 0.92 & 0.94 & 0.94 \\
\bottomrule

  \end{tabular}

  \end{table}


%

\begin{table}[h!]

  \centering

  \caption{Micro ${F}_{1}$ and Weighted ${F}_{1}$ scores of transformer models for sentiment analysis classification task using original dataset.} 

  \label{table:7}

  \begin{tabular}{lllll}

  \toprule
   & \multicolumn{2}{c}{Seed 0} & \multicolumn{2}{c}{Seed 42}\\

   \cmidrule(lr){2-3}\cmidrule(lr){4-5}

   Models & micro ${F}_{1}$ & weighted macro ${F}_{1}$ & micro ${F}_{1}$ & weighted macro ${F}_{1}$ \\

   \midrule
ALBERT base & 0.61 & 0.54 & 0.52 & 0.51 \\
ALBERT large & 0.52 & 0.43 & 0.59 & 0.54 \\
DistilBERT base cased & 0.56 & 0.55 & 0.67 & 0.66 \\
DistilBERT base uncased & 0.65 & \textbf{0.64} & 0.57 & 0.57 \\
DistilRoBERTa base & 0.62 & 0.60 & 0.68 & 0.67 \\
BERT base cased & 0.55 & 0.50 & 0.57 & 0.57 \\
BERT base uncased & 0.61 & 0.55 & 0.60 & 0.60 \\
BERT large cased & 0.57 & 0.58 & 0.61 & 0.60 \\
BERT large cased whole word masking & 0.47 & 0.46 & 0.66 & 0.64 \\
BERT large uncased & 0.66 & 0.59 & 0.57 & 0.57 \\
BERT large uncased whole word masking & 0.64 & 0.61 & 0.60 & 0.61 \\
RoBERTa base & 0.65 & 0.60 & 0.70 & \textbf{0.68} \\
RoBERTa large & 0.59 & 0.59 & 0.62 & 0.60 \\
XLNet base cased & 0.52 & 0.53 & 0.54 & 0.55 \\
XLNet large cased & 0.52 & 0.42 & 0.44 & 0.27 \\
\bottomrule

  \end{tabular}

  \end{table}
  
%


\begin{table*}
\centering
\caption{Examples of sentences that models correctly and incorrectly predicted the class label for topic-relevance classification task, True Label (TL) is a human label and Predicted Label (PL) is a model label.}
\label{table:9}
\begin{adjustbox}{width=\textwidth}
\begin{tabular}{p{0.1cm}p{8cm}p{0.3cm}p{0.3cm}p{8cm}p{0.3cm}p{0.3cm}}
\toprule
  & Correctly predicted labels & TL & PL & Incorrectly predicted labels & TL & PL \\
\midrule
1 & For instance, the Rook can be posted at a5 and prevent the Black King from attacking White's King's side pawns, whilst the White King makes for the R at h7 and effects its capture. & 1 & 1 & 
As soon as the Knight can obtain the King's support the game is drawn even when the King is already forced on to the edge of the board. & 1 & 0 \\

2 & Therefore White is condemned to inactivity. & 0 & 0 & 
Any other move We shall find that openings classed under C generally lead to positions treated under A and B. & 0 & 1 \\

3 & The game then would be drawn after 10. Kg2 Kf4 11. Kf2 because White maintains the opposition, and Black cannot get through at e3 or f3. & 1 & 1 & 
The Queen wins against any other piece; the Rook alone may give trouble. & 0 & 1 \\

4 & Therefore: in any combination which includes a number of exchanges on one square, all you have to do is to count the number of attacking and defending units, and to compare their relative values; the latter must never be forgotten. & 0 & 0 & 
If the pawns are still standing behind, the King who has the most advanced position has always the advantage, because he threatens to attack the opposing pawns should they leave their base. & 1 & 0 \\

5 & If Black were to play h4 at once, White would reply with 8. Kh3 and after hxg3 9. Kxg3. Black would have to give up the spare move c5 to gain the square at c4 for his King. & 1 & 1 & 
But a mate can be forced if the weaker side has a spare move which prevents the stalemate, e.g. Diagram 44. & 0 & 1 \\

6 & We shall refer to the execution of these plans later on. & 0 & 0 & 
It is a stalemate position. & 1 & 0 \\

7 & Black, however, must try to round off his pawn position on the Queen's side, by moving his c Pawn into line. & 1 & 1 & 
For the sake of completeness I will show a few cases in which Q or R cannot win against an advanced pawn. & 0 & 1 \\

8 & That is the ability to judge whether an end-game which could be brought about by exchanges is won or not; in other words, whether it can be reduced to one of the typical positions referred to above. & 0 & 0 & The ending of KING AND PAWN AGAINST KING is one of the simplest albeit one of the most important of elementary cases. 
& 0 & 1 \\

9 & The Queen wins against an advanced pawn, even when the latter is supported by the King; only the h3 or g4 pawn can draw sometimes, when the pawn is on the seventh supported by the King, and the opposing d7 cannot move to the queening square. & 1 & 1 & If now the White King comes up, he will in the end force the sacrifice of the Black Rook for the pawn, but meanwhile the Black King captures the White pawns, and with passed pawns on the King's side might get winning chances. 
& 1 & 0 \\

10 & The following example may serve as an illustration. & 0 & 0 & While White's development is easy and natural, Black has difficulty in finding good places for his King's side pieces. & 0 & 1 \\
\bottomrule
\end{tabular}
\end{adjustbox}
\end{table*}


\begin{table*}
\centering
\caption{Examples of sentences that models correctly and incorrectly predicted the class label for sentiment analysis classification task, True Label (TL) is a human label and Predicted Label (PL) is a model label.}
\label{table:10}
\begin{adjustbox}{width=\textwidth}
\begin{tabular}{p{0.1cm}p{8cm}p{0.3cm}p{0.3cm}p{8cm}p{0.3cm}p{0.3cm}}
\toprule
  & Correctly predicted labels & TL & PL & Incorrectly predicted labels & TL & PL \\
\midrule
1 & But in any case there remains the disadvantage that White was forced to play the c Pawn, whilst before he had the option of withholding its advance until a more opportune moment. & 0 & 0 & 

In Diagram 81, for instance, White is lost, as Black gives up his Rook at d2 and plays g3, after which one of the pawns queens. & 2 & 0 \\

2 & 1. Ka7 Qa4+ 2. Kb6 Qb4+ 3. Kc7 Qc5+ 4. Kd8 Qd6+ 5. Kc8 Qc6+ 6. Kb8 Kc5 7. Ka7 Qa4+ 8. Kb8 Kc6 9. Kc8 Qa6 etc. & 1 & 1 & 
The ensuing end-game, however, is inferior for Black, because the d Pawn is weak and White threatens eventually to force his Queen's Pawn through. & 3 & 0 \\

3 & In eleven moves Black captures the opposing c Pawn and queens his own. & 2 & 2 & 
Black might be justified in taking the pawn, if he really could hold the pawn thus gained. & 1 & 2 \\

4 & Therefore he prepares the manoeuvre Nf3, Nd2, Nc4. & 1 & 1 & 
Whilst White has a pawn firmly posted in the centre, Black has a Knight there which will soon be driven away. & 3 & 2 \\

5 & White has the advantage, because Black must keep either his King or his Knight permanently near the passed pawn, guarding against its advance, whilst both White's King and Knight can attack the Black pawns. & 2 & 2 & 
In the position in Diagram 53 Black plays Kd5 and maintains the opposition until the pawn moves, after which a typical position, similar to the one treated in connection with Diagram 50 is brought about. & 1 & 2 \\

6 & But he cannot maintain it after Black's Kh3 because the square at d6 for distant opposition is not accessible. & 0 & 0 & 
From these considerations the following development seems to be natural: 8. c3 Na5 9. Bc2 c5 10. d4 Qc7 (to support the e Pawn ) it leads to the position in Diagram 23. & 2 & 1 \\

7 & Black, however, must try to round off his pawn position on the Queen's side, by moving his c Pawn into line. & 2 & 2 & 
But playing into the third rank is of no use, as White then assumes the direct opposition, and wins as in Diagram 60. & 3 & 0 \\

8 & A better plan would be 6. f3, preventing Ne4 and preparing the eventual advance of the King's Pawn to e4. & 2 & 2 & 
From these considerations the following development seems to be natural: 8. c3 Na5 9. Bc2 c5 10. d4 Qc7 (to support the PE it leads to the position in Diagram 23. & 2 & 1 \\

9 & If, on the other hand, the Black King tries to obstruct the way to the Queen's side, White penetrates into the Black pawn position. & 3 & 3 & Quite a different system of opening ensues, when Black does not delay pushing the P to c5 until after his pieces are developed, but makes the advance on his third move. & 1 & 2 \\

10 & It can be easily perceived that the Bishop cannot capture the pawn at c7 on account of a6. & 0 & 0 & If the Black King comes to the rescue of the Bishop, the other pawn proves Black's downfall. & 0 & 3 \\

\bottomrule
\end{tabular}
\end{adjustbox}
\end{table*}

\begin{table*}
\centering
\caption{Examples of synthetic sentences after manually checked the labels for sentiment analysis classification task. Original Label (OL) is the original sentence label automatically assigned to the corresponding synthetically generated sentence, (TL) manually label given after checking the context of the synthetically generated sentence.}
\label{table:11}
\begin{adjustbox}{width=\textwidth}
\begin{tabular}{p{0.1cm}p{8cm}p{8cm}p{0.3cm}p{0.3cm}}
\toprule
  & Original sentence & Synthetic sentence & OL & TL  \\
\midrule
1 & On principle it does not seem advisable for Black to play d5 on the first move in reply to 1. e4. Although White's centre pawn disappears after 2. exd5 Qxd5 Black loses a move through 3. Nc6 and his Queen has no place from which it cannot be driven away very soon, unless it be at d8. & 

On principle it does not seem advisable for Black to play 2. Qd2 in response to 1. e4. Although White's centre pawn disappears after 1. d3 Nc6 2. Qd2 Qxd5 Black loses a move through 3. Nc6 and his Queen has no place from which it cannot be driven away very soon, unless it be at d8. & 0 & 0 \\

2 & On the other hand, it cannot be denied that Black's pawn formation on the d side is weak, and that his centre is less secure.  & 
I think this is true. But the fact that Black has a weakened centre does not automatically imply that Black is weaker. In fact, Black has a very solid centre, with two pieces that are very hard to get rid of, a rook and a bishop. & 0 & 2 \\

3 & 5. Bxf6 followed by Qh4+) 5. f3. & 
5. Qh4+ followed by f3 followed by Qxf6 followed by Qh5+ followed by Bxf3. 5. f3. & 1 & 1 \\

4 & Diagram 26 shows the position after 9. c3 & 
The position after 8. d4 is not clear. I think 8...Nf6 was a mistake, since the knight can now be exchanged. & 1 & 0 \\

5 & White can confine himself to a simple development such as: Nf3, Bb5, e3, Nd2 (Nc3 would only be good if preceded by c4, because Black would again lead into a Queen's gambit with d5 and c4). & 
White can also use the Queen's Gambit with d5 and e3, but only in the position shown above. & 2 & 2 \\

6 & Black's pawns at e5 and c5 then exert a pressure on White's d5. & 
Black's queen is in the center of the board and the queen is threatening to capture Black's bishop. Black's bishop is at a4 and his queen at a5. The bishop at e5 is a weak point of the game. The pawn structure is a bit strange, as Black's pawns are at a5 and f6. & 2 & 3 \\

7 & Here the possession of an extra pawn is seldom sufficient for a win, unless the stronger side has also an advantage in the greater mobility of the Rook. & 
The advantage of the Rook in mobility is not always a good one. & 3 & 0 \\

8 & If Black now enters his second rank, White immediately plays into opposition on his third rank, e.g. 1. ... Kd7 2. Kd3 and still maintains it by 3. Ke3 if Black plays a waiting move such as 2. ... Ke7. Now Black has no further waiting moves, as White threatens to capture one of the pawns. & 
If White plays 2.... Kd7, Black plays 3. Ke3 and White cannot prevent it. & 3 & 1 \\

9 & He can only attack the pawn from g7 or g8, both of which are inaccessible as the Black King gets to g8. & If he attacks the rook from g7, the Black King will have to retreat and the pawn will be safe. If he attacks the pawn from g8, the Black King will be able to attack the rook from g8, and the White King will be safe. & 3 & 3 \\

10 & No fault can be found with this move, although it blocks the Bishop, since the latter can only be developed effectively at c2. & The move is a very good one for White, since it blocks the Bishop and forces Black to develop his Queen to c3. & 1 & 2 \\

\bottomrule
\end{tabular}
\end{adjustbox}
\end{table*}


\end{document}